\documentclass[aip,amsmath,amssymb,reprint,superscriptaddress,floatfix]{revtex4-1}

\usepackage[utf8]{inputenc}
\usepackage[T1]{fontenc}
\usepackage{amsmath,amssymb,amsfonts}
\usepackage{graphicx}
\usepackage{booktabs}
\usepackage{multirow}
\usepackage{xcolor}
\usepackage{hyperref}
\usepackage{cleveref}
\usepackage{enumitem}
\usepackage{tikz}
\usetikzlibrary{shapes.geometric, arrows.meta, positioning, calc}
\usepackage{xspace}

\newcommand{\benchname}{TPS-CalcBench\xspace}
\newcommand{\vprimary}{v4(420)\xspace}
\newcommand{\vnoisy}{v2(810)\xspace}

\begin{document}

\title{\benchname: A Benchmark and Diagnostic Evaluation Framework for LLM Analytical Calculation Competence in Hypersonic Thermal Protection System Engineering}

\author{ZHENG Jinglai}
\affiliation{School of Civil Engineering, Beijing Jiaotong University, Beijing 100044, China}
\author{QIAO Chuhan}
\affiliation{School of Civil Engineering, Beijing Jiaotong University, Beijing 100044, China}
\author{HUANG Haiming}
\email{24120979@bjtu.edu.cn}
\affiliation{School of Civil Engineering, Beijing Jiaotong University, Beijing 100044, China}

\date{}  

\begin{abstract}
Deploying large language models (LLMs) as reasoning assistants in safety-critical aerospace engineering demands a qualitatively different standard of evaluation than general scientific benchmarks provide. In hypersonic thermal protection system (TPS) design, a single miscalculated stagnation-point heat flux or erroneous boundary-layer estimate can propagate silently into a catastrophic design margin violation---making a model that produces a \emph{numerically plausible yet physically unjustified} answer more dangerous than one that refuses to answer. Yet no existing benchmark probes this failure mode: current scientific reasoning suites test either abstract mathematics or introductory physics, and evaluate only final-answer correctness, leaving the process of engineering reasoning entirely unscrutinized.

We present \benchname, a benchmark and diagnostic evaluation framework for \emph{analytical} (closed-form, formula-derivable) calculation tasks in hypersonic aerodynamics and high-temperature gas dynamics. To the best of our knowledge, this is the first benchmark specifically targeting this class of calculations that experienced TPS engineers perform without simulation tools. Our work makes five contributions. \textbf{(1) Domain-grounded task taxonomy.} We define a rigorous scope covering four difficulty levels (L1--L4) and eight domain categories grounded in Anderson's authoritative graduate textbook~\citep{anderson2006hypersonic}, deliberately isolating analytical tasks from simulation-dependent problems. \textbf{(2) Dual-track evaluation framework.} We design complementary measurement tracks for outcome correctness and reasoning process quality as independently measurable axes: a relative-error outcome scorer with unit verification, and an 8-dimension rubric scored by a calibrated LLM judge with expanded human audit ($n=62$ calibration items)---enabling detection of the ``right answer, wrong reasoning'' failure mode invisible to answer-only evaluators. \textbf{(3) Trustworthiness-engineered data pipeline.} We develop a human--AI collaborative construction pipeline distilling 4,560 raw candidates into a 420-item high-confidence core set (v4), retaining an 810-item pre-gating set (v2) as a noise-injection control. \textbf{(4) Empirical noise-sensitivity analysis.} We directly measure the effect of benchmark data quality on model rankings and KPI estimates by running controlled experiments on both the curated and pre-gating sets. \textbf{(5) Diagnostic-informed pilot interventions.} We derive three improvement strategies from the rubric failure taxonomy as proof-of-concept mitigation pilots: Domain-Formula Alignment fine-tuning (DFA-TPS), Retrieval-Augmented Equation Grounding (RAG-EQ), and Process-Aware Chain-of-Thought (PA-CoT) prompting. Experiments across 13 models from 7 families reveal a wide competence spread (KPI 12.6--87.9), systematic formula-selection deficiencies invisible to outcome-only evaluation, data-quality-induced rank reshuffling, and promising gains from targeted diagnostic interventions---establishing a practical \emph{diagnose $\to$ evaluate $\to$ intervene} framework for LLM deployment assessment in safety-critical engineering.

\end{abstract}

\maketitle

\section{Introduction}
\label{sec:introduction}

The prospect of using large language models (LLMs) as automated reasoning assistants in aerospace engineering is both compelling and sobering~\citep{openai2023gpt4, anthropic2024claude, google2024gemini}. Compelling, because LLMs can potentially accelerate preliminary design, automate sanity checks, and democratize access to specialized domain knowledge. Sobering, because the failure modes of LLMs in high-stakes engineering settings are qualitatively different---and more treacherous---than those observed in general-purpose benchmarks. But sobering does not mean intractable: if we can precisely characterize \emph{where} and \emph{why} models fail, we can design targeted interventions. This paper presents \benchname as an end-to-end framework for doing exactly that---building a rigorous benchmark, diagnosing model deficiencies with engineering-grounded rubrics, and deriving concrete improvement strategies from the diagnostic evidence.

\paragraph{The evaluation gap in professional engineering calculations.}
The benchmark landscape for scientific reasoning has grown rapidly~\citep{cobbe2021gsm8k, hendrycks2021math, sun2024scieval, wang2023scibench}. Yet existing evaluations share a common architectural limitation: they measure whether a model produces the \emph{correct final answer}, not whether it arrived there through \emph{physically valid reasoning}. For introductory mathematics or general science, this conflation is tolerable---a lucky numerical coincidence is unlikely and harmless. For professional engineering calculations, it is neither. An engineer who trusts a model's stagnation-point heat-flux estimate without scrutinizing the formula selection and regime assumptions is exposed to a systematically dangerous failure mode: the model's answer is numerically plausible but physically unjustified, and no alarm bell rings.

\paragraph{Why TPS calculations are uniquely challenging.}
Thermal protection system (TPS) design for hypersonic vehicles sits at the intersection of several cognitively demanding disciplines. Computing stagnation-point heat flux via the Fay--Riddell correlation, for instance, requires the analyst to: (i) identify the applicable flow regime ($M_\infty > 5$, viscous-interaction-dominated), (ii) select the correct boundary condition formulation (catalytic vs.\ non-catalytic wall), (iii) correctly apply the Lewis number correction for chemically reacting boundary layers, (iv) maintain SI unit consistency across a seven-parameter expression, and (v) verify that the result is physically plausible given the altitude--velocity combination. Missing any one of these steps produces a wrong answer that may \emph{appear} correct to an evaluator checking only the final number. The same multi-step, regime-sensitive, unit-consistent reasoning structure applies to calculations of boundary-layer transition, oblique shock thermodynamics, equilibrium chemical composition, vibrational nonequilibrium, and radiative transfer---making TPS design a natural stress test for LLM physical reasoning capability.

\paragraph{A dual-axis diagnostic imperative.}
We argue that evaluating LLMs on professional engineering calculations requires two \emph{orthogonal} diagnostic axes: (1) \textbf{outcome correctness}---does the predicted numerical value agree with the reference?---and (2) \textbf{process trustworthiness}---does the solution path reflect physically valid, domain-appropriate reasoning? These axes are logically independent: a model can achieve outcome correctness through compensating errors, memorized numerical patterns, or lucky dimensional analysis, while a model with transparent, well-reasoned derivations may make a minor arithmetic slip. An evaluation framework that collapses these two axes into a single scalar conflates competence with luck and conceals the failure modes that matter most in engineering contexts.

\paragraph{The benchmark construction challenge.}
Building a domain-specific benchmark that is simultaneously high-coverage, high-fidelity, and reliably scored is itself a non-trivial research problem. Extracting calculation problems from graduate engineering textbooks introduces a cascade of quality hazards: OCR artifacts, figure-fragment misidentifications, incomplete given-parameter sets, fallback target specifications, and reference solutions that presuppose simulation infrastructure. Addressing these hazards with ad-hoc filtering produces benchmarks that are technically usable but whose quality characteristics are opaque, making it impossible to know whether benchmark differences reflect model capability or data quality variance.

\subsection{Contributions}

This paper addresses the above challenges through five principled contributions.

\paragraph{Contribution 1: TPS Analytical Calculation Benchmark.}
We present \benchname, a benchmark targeting \emph{analytical} (closed-form, formula-derivable) calculation tasks in hypersonic aerodynamics and high-temperature gas dynamics---the class of calculations that experienced TPS engineers perform for preliminary design, physical sanity checking, and regime identification, without recourse to CFD or FEA simulation. To the best of our knowledge, no prior benchmark specifically scopes this class of problems at the professional engineering level. The benchmark spans four difficulty levels (L1: single-step substitution through L4: coupled iterative reasoning) and eight domain categories (Newtonian aerodynamics, shock relations, boundary-layer theory, aerothermal heating, viscous interaction, chemical equilibrium, nonequilibrium flow, and radiation), sourced systematically from Anderson's \emph{Hypersonic and High-Temperature Gas Dynamics}~\citep{anderson2006hypersonic}. By isolating the analytical scope, we enable infrastructure-free evaluation of physical reasoning and multi-step calculation competence.

\paragraph{Contribution 2: Dual-Track Engineering Evaluation Protocol.}
We design a two-track evaluation architecture that treats outcome correctness and reasoning process quality as independently measurable quantities. Track 1 uses relative-error banding with explicit unit verification, distinguishing four correctness levels. Track 2 employs an 8-dimension rubric aligned with the verification steps a competent TPS engineer would apply when reviewing a junior colleague's calculation: formula selection, parameter identification, dimensional consistency, arithmetic accuracy, physical plausibility, assumption transparency, result interpretation, and presentation quality. The rubric is scored by a LLM judge calibrated against domain-expert human ratings, enabling scalable yet principled process-level assessment. This dual-axis design exposes failure modes---including ``right answer via wrong formula,'' ``plausible answer via undisclosed assumption,'' and ``domain hallucination''---that are structurally invisible to answer-only evaluators.

\paragraph{Contribution 3: Trustworthiness-Engineered Data Construction Pipeline.}
We develop a rigorous, auditable pipeline for constructing domain-specific benchmarks from textbook sources, organized around the principle that \emph{benchmark conclusions are only as credible as their underlying data}. The pipeline combines high-recall automated extraction (dual rule-based + LLM-assisted passes), five-stage progressive filtering (deduplication, rule-based problem validation, automated QC, human review and adjudication, and formal go/no-go gating), triple-reviewer checkpoints, targeted repair sprints for salvageable items, and version-locked dataset freezing with full provenance metadata. This pipeline reduces 4,560 raw candidates to a 420-item high-confidence core set verified to meet strict standards of self-containedness, solution verifiability, and scoring robustness. The pipeline design is explicitly transferable to other domain-specific benchmarks constructed from technical literature.

\paragraph{Contribution 4: Empirical Benchmark Quality Sensitivity Analysis.}
Rather than asserting that data quality matters, we \emph{measure} it. By maintaining the pre-gating 810-item set (v2) as an explicit noise-injection control alongside the curated 420-item set (v4), and running identical experiments on both, we directly quantify the effect of benchmark noise on model rankings, KPI magnitudes, and evaluation conclusions. This contribution provides the field with concrete empirical evidence for quality-gating investment---addressing a concern widely acknowledged in the benchmark construction literature~\citep{northcutt2021pervasive, biderman2024lessons} but rarely addressed with controlled experimental evidence.

\paragraph{Contribution 5: Diagnostic-Informed Pilot Interventions.}
A benchmark is only instrumentally valuable if its diagnostic output translates into actionable improvement signals. We close the diagnostic-to-intervention loop by deriving three pilot mitigation strategies from the rubric failure taxonomy. \textbf{Strategy I: DFA-TPS (Domain-Formula Alignment Fine-Tuning)} constructs a domain-specific SFT dataset targeting G1 (formula selection) deficits that account for the largest share of process-score failures. \textbf{Strategy II: RAG-EQ (Retrieval-Augmented Equation Grounding)} augments model generation with a curated 847-equation knowledge base (EKB) and a regime-aware retriever to suppress hallucinated correlations. \textbf{Strategy III: PA-CoT (Process-Aware Chain-of-Thought)} instantiates a 7-step Engineering Calculation Protocol as a structured reasoning scaffold, enforcing multi-step discipline that separates expert engineering reasoning from numerical pattern matching. These interventions are benchmark-informed pilot studies; their transferability beyond the present benchmark setting remains to be validated and is discussed in Section~\ref{sec:limitations}.

\section{Related Work}
\label{sec:related_work}

\paragraph{Mathematical and Scientific Reasoning Benchmarks.}
Benchmarks for LLM reasoning span a wide capability range: from arithmetic word problems (GSM8K~\citep{cobbe2021gsm8k}) to competition-level mathematics (MATH~\citep{hendrycks2021math}, AIME~\citep{aime2024}) and multi-domain science (MMLU~\citep{hendrycks2021mmlu}, SciEval~\citep{sun2024scieval}). The closest prior work to ours is SciBench~\citep{wang2023scibench}, which evaluates LLMs on college-level physics and chemistry problems from undergraduate textbooks and includes free-form solution extraction. However, SciBench's coverage stops at introductory thermodynamics and classical mechanics---it does not address graduate-level engineering correlations, hypersonic regime-sensitivity, or the multi-domain knowledge integration (e.g., coupling shock relations with boundary-layer theory and real-gas thermodynamics) that characterizes TPS analytical work. Critically, SciBench, like all benchmarks in this category, evaluates only final-answer correctness, leaving the process of engineering reasoning entirely unmeasured.

\paragraph{Domain-Specific Engineering Benchmarks.}
Efforts to evaluate LLMs on engineering problems have produced benchmarks for structural analysis, circuit design, and fluid mechanics~\citep{zhang2024cfdgpt}. These benchmarks, however, share two limitations relevant to our setting: they either require numerical simulation infrastructure (CFD solvers, FEA packages) as part of the task, or they focus on qualitative understanding and concept identification rather than quantitative calculation. \benchname occupies an orthogonal niche: it targets the class of problems that experienced engineers solve analytically---without simulation tools---as part of preliminary design, physical intuition building, and sanity-checking. This ``analytical regime'' is a necessary prerequisite for simulation-augmented work and a distinct test of domain reasoning capability that has not been benchmarked before.

\paragraph{Process-Level and Verifiable Reasoning Evaluation.}
The inadequacy of outcome-only evaluation has been recognized in the mathematical reasoning literature, motivating process reward models (PRMs)~\citep{lightman2023prm, uesato2022process} that score individual reasoning steps. Our approach differs in two important ways. First, we do not train a step-level verifier; instead, we define \emph{domain-grounded rubric dimensions} that map to the specific verification actions a TPS engineer would apply. Second, our rubric captures engineering-specific quality properties---physical plausibility, regime appropriateness, unit consistency across the full solution chain---that generic PRMs do not model. The result is a process evaluation instrument calibrated to the epistemology of the domain, not just the logic of mathematical derivation.

\paragraph{Benchmark Construction and Data Quality.}
The importance of rigorous benchmark construction has been underscored by evidence of pervasive label errors~\citep{northcutt2021pervasive}, contamination artifacts~\citep{yang2023contamination}, and the sensitivity of evaluation conclusions to annotation quality~\citep{biderman2024lessons}. Most benchmark papers acknowledge these concerns in limitations sections but provide no controlled evidence for their magnitude. Our noise-sensitivity experiment is a direct response: by maintaining both a rigorously curated set and a noisier pre-gating set, we quantify---rather than merely assert---the effect of data quality on model rankings.

\paragraph{Positioning Summary.}
\Cref{tab:benchmark_positioning} summarizes the key distinctions between \benchname and representative prior benchmarks across the five dimensions most relevant to our contributions.

\begin{table*}[htbp]
\centering
\caption{Positioning of \benchname against representative benchmarks. \checkmark = fully supported; $\sim$ = partial; $\times$ = absent.}
\label{tab:benchmark_positioning}
\small
\setlength{\tabcolsep}{3pt}
\begin{tabular}{@{}lccccc@{}}
\toprule
\textbf{Benchmark} & \textbf{Prof.\ Eng.} & \textbf{Multi-step} & \textbf{Process} & \textbf{Unit} & \textbf{Quality} \\
 & \textbf{domain} & \textbf{calc.} & \textbf{eval.} & \textbf{verif.} & \textbf{sensitivity} \\
\midrule
GSM8K & $\times$ & $\times$ & $\times$ & $\times$ & $\times$ \\
MATH & $\times$ & \checkmark & $\times$ & $\times$ & $\times$ \\
SciBench & $\sim$ & $\sim$ & $\times$ & $\sim$ & $\times$ \\
CFD-related & \checkmark & $\times$ & $\times$ & $\times$ & $\times$ \\
\benchname (ours) & \checkmark & \checkmark & \checkmark & \checkmark & \checkmark \\
\bottomrule
\end{tabular}
\end{table*}

\section{Benchmark Construction}
\label{sec:construction}

Constructing a domain-specific engineering benchmark from graduate textbook sources is not merely a data engineering task---it is a \emph{trustworthiness engineering} challenge. A benchmark item is trustworthy if and only if it satisfies three jointly necessary properties: (P1) \textbf{self-containedness} (the problem is solvable from the stated given parameters alone, without implicit reference to figures, tables, or solution context); (P2) \textbf{solution verifiability} (a unique, numerically definite reference answer exists and has been expert-confirmed); and (P3) \textbf{scoring robustness} (the evaluation protocol can reliably distinguish a correct from an incorrect model response). Violating any one property makes benchmark conclusions unreliable in a way that is invisible to downstream consumers.

\begin{figure*}[t]
\centering
\includegraphics[width=\textwidth]{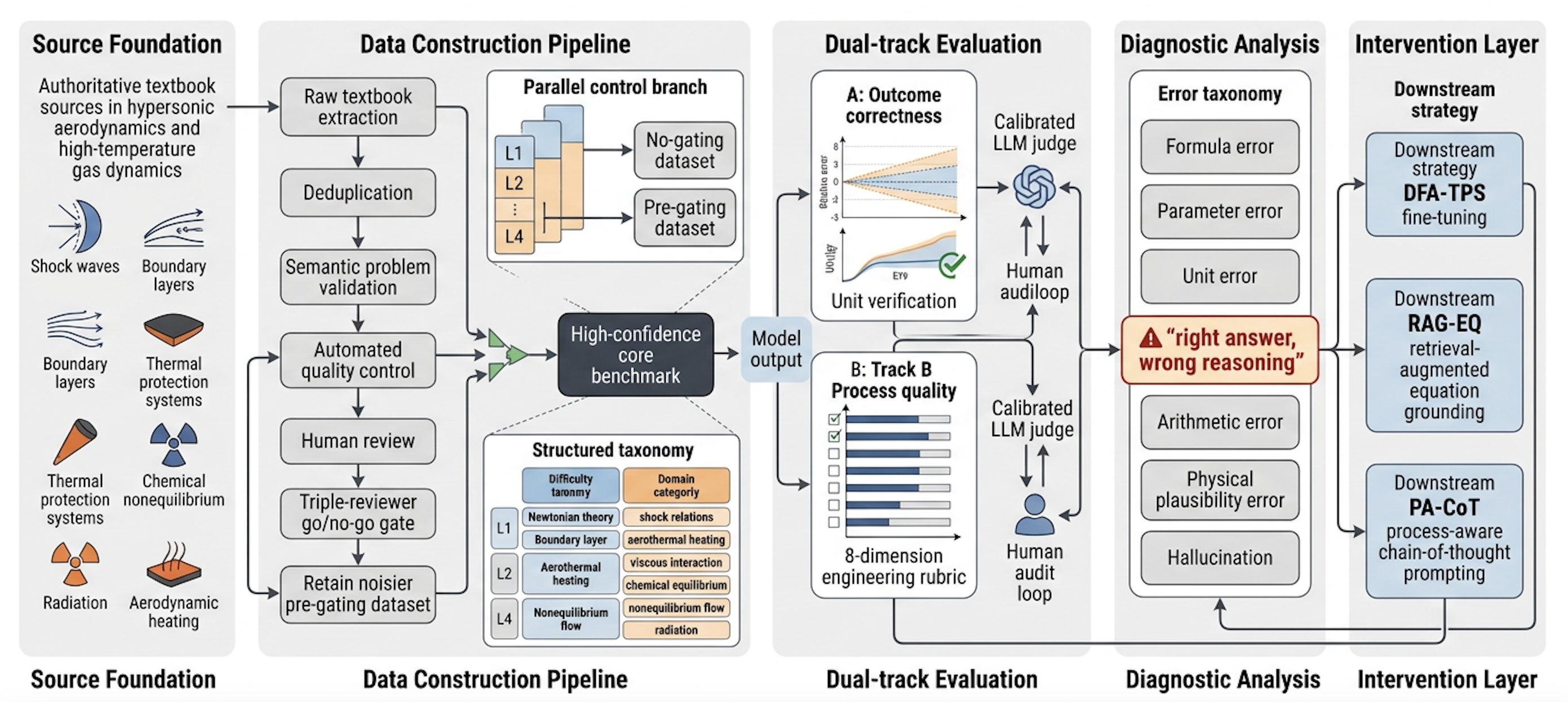}
\caption{Data construction funnel for \benchname. The pipeline proceeds from high-recall automatic extraction to increasingly stringent quality gates, culminating in a triple-reviewer adjudication. The final core set (v4) contains 420 fully verified items; the intermediate v2 set (810 items) serves as a noisier comparison baseline.}
\label{fig:funnel}
\end{figure*}

The central challenge is that automatic extraction from technical documents introduces systematic violations of all three properties: OCR artifacts corrupt numerical parameters (P1), equation-reference misparses produce problems with no deterministic answer (P2), and fallback target names produce unevaluable items (P3). Our pipeline is specifically designed to detect and remediate these failure modes at scale, applying progressively more expensive quality gates as the candidate pool narrows. \Cref{fig:funnel} illustrates the complete construction pipeline with item counts at each stage.

\subsection{Data Sources and Scope}
\label{sec:data_sources}

All problems are derived from Anderson's \emph{Hypersonic and High-Temperature Gas Dynamics} (2nd Edition)~\citep{anderson2006hypersonic}, a standard graduate-level textbook covering hypersonic aerodynamics, boundary-layer theory, aerothermal heating, chemical equilibrium, nonequilibrium flow, radiation gas dynamics, and related topics. We restrict scope to \emph{analytical} tasks: problems solvable via closed-form formulas, engineering correlations, or standard thermodynamic/aerodynamic relations, without requiring CFD, FEA, or Monte Carlo simulation. This restriction serves two purposes: it isolates the LLM's reasoning and calculation capability from tool-use infrastructure, and it ensures that reference solutions are fully verifiable without specialized software.

\subsection{Extraction Strategy and Scope Isolation}
\label{sec:extraction}

\paragraph{High-recall dual extraction.}
We deliberately begin with a permissive extraction strategy (AUTO\_EXTRACT\_RELAXED) that prioritizes recall over precision. A rule-based pass scans for segments containing calculation-imperative verbs (``calculate,'' ``determine,'' ``find,'' and their common Chinese-language counterparts, here referenced via the transliterations ``jisuan,'' ``qiu,'' and ``queding''), numerical parameters with units, and equation references. An LLM-assisted second pass identifies structurally valid calculation tasks that lack explicit imperative markers but contain sufficient given--target structure for analytical solution. Crucially, both passes operate independently, and their union forms the initial candidate pool. This aggressive dual-pass strategy is intentional: false negatives (missing genuine problems) are unrecoverable at later stages, whereas false positives (spurious candidates) are recoverable through downstream filtering. The result is 4,560 initial candidates with an estimated recall above 95\% and a precision around 25\%.

\paragraph{Analytical scope isolation.}
A defining design choice of \benchname is the explicit exclusion of simulation-dependent problems. Table~\ref{tab:scope} formally defines the inclusion and exclusion boundary, resolving edge cases that arise frequently in TPS literature.

\begin{table*}[t]
\centering
\small
\caption{Analytical scope boundary for \benchname. The ``Included'' column defines the accepted problem class; ``Excluded'' lists rejected types with representative examples.}
\label{tab:scope}
\begin{tabular}{@{}p{\dimexpr0.44\columnwidth\relax}p{\dimexpr0.44\columnwidth\relax}@{}}
\toprule
\textbf{Included} & \textbf{Excluded} \\
\midrule
Direct formula substitution (single or multi-step) & CFD/FEA solver setup or post-processing \\
Textbook engineering correlations (Fay--Riddell, Eckert, etc.) & Monte Carlo simulation results \\
Algebraic derivation of classical results & Numerical PDE integration beyond hand-iteration \\
Standard thermodynamic state relations (Rankine--Hugoniot, etc.) & Machine learning or data-driven prediction tasks \\
Low-order iterative convergence (max 3--5 steps) & Multi-physics coupling loop execution \\
Equilibrium composition via tabulated or algebraic models & Problems requiring figure-reading without stated values \\
Radiative transfer with closed-form approximations (optically thin/thick limits) & Problems with reference solutions requiring simulation software \\
\bottomrule
\end{tabular}
\end{table*}

Edge cases: equilibrium composition problems are included if solvable by algebraic manipulation of equilibrium constants; they are excluded if they require numerical solution of a system of nonlinear equations without closed-form reduction. Viscous interaction parameter calculations are included because they reduce to algebraic expressions; full viscous/inviscid interaction solutions requiring iterative field convergence are excluded. The acceptance/rejection criterion was documented for all 770 Stage 4 rejects to enable auditable review.

This isolation serves two purposes. First, it creates a well-defined capability boundary: mastery of the analytical scope is a necessary (though not sufficient) precondition for simulation-augmented work. Second, it ensures reference solutions are fully verifiable by domain experts without specialized software.

\subsection{Data Schema}
\label{sec:schema}

Each benchmark item is represented as a structured JSON object with the following fields: a unique identifier (\texttt{id}), difficulty level (\texttt{level}: L1--L4), task type (\texttt{task\_type}: \texttt{numerical\_calc}, \texttt{derivation}, \texttt{comparison}), domain tags (\texttt{domains}), the complete problem statement (\texttt{question}), a list of given parameters (\texttt{given}, each with name, value, unit, and description), a list of target quantities (\texttt{targets}, each with name, expected unit, scoring weight, and description), and metadata tracing provenance to the source chapter and extraction method. The schema enforces self-containedness: each item must be solvable from its \texttt{given} fields alone, without reference to external figures, tables, or preceding examples.

\subsection{Multi-Stage Quality Gating Pipeline}
\label{sec:quality_gating}

The 4,560 raw candidates undergo a five-stage progressive quality gate designed to enforce Properties P1--P3 at increasing stringency, with the most expensive (human) verification applied only to candidates that have survived all automated gates. This design minimizes human annotation cost while maximizing the probability that retained items are genuinely trustworthy. \Cref{tab:pipeline_stages} summarizes each stage's purpose, mechanism, and yield.

\paragraph{Stage 1: Deduplication and OCR Remediation.}
The first gate targets two corruption classes that are universal in textbook extraction: near-duplicate items arising from repeated chapter examples, and OCR artifacts that corrupt numerical parameters. Near-duplicates are identified by Jaccard similarity on tokenized question text (threshold $\tau=0.85$) and merged with provenance tracking. OCR artifacts---garbled Unicode sequences, concatenated page numbers, misrecognized mathematical symbols (e.g., $\beta$ rendered as ``{3'')---are corrected using a rule-based normalization dictionary derived from inspection of the initial candidate pool. This stage reduces the pool from 4,560 to approximately 2,200, primarily by eliminating duplicates that arise from the textbook's worked-example repetition structure.

\paragraph{Stage 2: Semantic Problem Validation.}
A battery of rule-based semantic checks eliminates candidates that are structurally not calculation problems. Three patterns account for the majority of false positives from the extraction stage: (i)~\emph{narrative passages}---textbook discussion paragraphs that contain numerical values but no imperative calculation request; (ii)~\emph{figure and table fragments}---axis labels, coordinate sequences, and legend text misidentified as given-parameter lists, detectable by the presence of $\geq6$ concatenated digits or axis-label patterns (\texttt{x/cm}, \texttt{y/m}); and (iii)~\emph{equation-reference misparses}---items whose ``given'' fields consist entirely of equation-number tokens (e.g., $\{6, 85\}$ from ``Eq.~(6-85)'') rather than physical parameters. A small number of manually identified problems missed by both extraction passes are also added at this stage. The resulting set of 2,250 candidates has been confirmed to contain a genuine calculation task structure, though completeness and correctness are not yet verified.

\paragraph{Stage 3: Automated Scoring-Feasibility QC.}
Property P3 (scoring robustness) is verified automatically before investing human review effort. Each candidate is checked for: (a)~\emph{target non-fallback completeness}---at least one target must have an explicitly named quantity with a physically resolvable unit, not a generic fallback label; (b)~\emph{given-parameter dimensional sufficiency}---the declared given parameters must span the dimensional basis required to compute the stated targets; and (c)~\emph{level--complexity consistency}---the assigned difficulty level must be coherent with the estimated number of reasoning steps in a reference solution. Items failing any check are flagged \texttt{warn} (borderline, forwarded to human review with annotations) or \texttt{reject} (unsuitable, removed). This stage passes 1,930 candidates while surfacing 320 items that would have been impossible to score reliably.

\paragraph{Stage 4: Domain-Expert Human Review and Adjudication.}
Surviving candidates are packaged into structured review batches and evaluated by domain-expert reviewers against four criteria that require human physical judgment: (i) self-containedness and absence of implicit context dependencies; (ii) completeness of the given-parameter list for a unique analytical solution; (iii) well-definedness and unambiguity of the target quantities; and (iv) correctness of the reference solution. Each item receives a disposition of \texttt{accept}, \texttt{revise} (actionable repair identified, queued for the repair sprint), or \texttt{reject}. This stage is the primary enforcer of Properties P1 and P2, reducing the pool to 1,160 accepted or revision-queued items while eliminating the 770 candidates that fail human-level scrutiny despite passing automated checks.

\paragraph{Stage 5: Triple-Reviewer Go/No-Go Gating and Targeted Repair Sprint.}
The final gate applies a formal \emph{triple-reviewer go/no-go protocol} that evaluates dataset readiness across three independent dimensions: \textbf{data quality} (field completeness, solution verifiability, self-containedness rate), \textbf{evaluation feasibility} (answer extractability from model outputs, unit normalization coverage), and \textbf{experimental reproducibility} (split-manifest stability, pipeline script executability, random-seed documentation). All three reviewers must independently reach \textsc{go} status on all three dimensions for the dataset version to be cleared for experimentation---a unanimity requirement that prevents premature release driven by any single reviewer's optimism.

The initial go/no-go checkpoint revealed critical deficiencies in the revision queue: a cluster of items with structurally incomplete given-parameter sets that required expert reconstruction rather than minor copy-editing. Rather than expanding the repair budget indefinitely, we applied a \emph{minimal repair sprint}---a time-boxed, effort-bounded remediation pass targeting only the highest-yield salvageable items. After repair and re-gating, the final core set (\vprimary) contains 420 items that have passed all five stages and received explicit triple-reviewer clearance. The pre-sprint set (\vnoisy, 810 items) is deliberately retained as a noise-injection control for the sensitivity analysis in \Cref{sec:noise_sensitivity}.

\begin{table*}[t]
\centering
\caption{Pipeline stage summary. Each stage enforces a subset of the trustworthiness properties P1--P3. ``Cost'' reflects the dominant resource type required.}
\label{tab:pipeline_stages}
\small
\begin{tabular}{@{}lllcc@{}}
\toprule
\textbf{Stage} & \textbf{Primary mechanism} & \textbf{Properties enforced} & \textbf{Output} & \textbf{Cost} \\
\midrule
S1: Dedup + OCR & Jaccard similarity, norm. dict. & P1 (partial) & $\sim$2,200 & Compute \\
S2: Semantic validation & Rule-based pattern matching & P1, P2 (partial) & 2,250 & Compute \\
S3: Automated QC & Dimensional + label checks & P3 & 1,930 & Compute \\
S4: Human review & Expert adjudication & P1, P2 (full) & 1,160 & Human \\
S5: Go/no-go + repair & Triple-reviewer gate & P1, P2, P3 (full) & \textbf{420} & Human \\
\bottomrule
\end{tabular}
\end{table*}

\subsection{Task Taxonomy}
\label{sec:taxonomy}

\benchname items span four difficulty levels and eight domain categories, as summarized in \Cref{tab:taxonomy}.

\begin{table*}[t]
\centering
\caption{Task taxonomy of \benchname. Difficulty levels are defined by the number of reasoning steps and prerequisite knowledge depth.}
\label{tab:taxonomy}
\small
\begin{tabular}{@{}llp{7cm}c@{}}
\toprule
\textbf{Level} & \textbf{Label} & \textbf{Definition} & \textbf{Count (v4)} \\
\midrule
L1 & Single-step & Direct formula substitution with given parameters & 170 \\
L2 & Multi-step & Chained calculations requiring 2--4 intermediate results & 120 \\
L3 & Cross-domain & Requires combining results from multiple physical domains & 80 \\
L4 & Advanced & Requires iterative or coupled solution strategies & 50 \\
\midrule
\multicolumn{3}{r}{\textbf{Total}} & \textbf{420} \\
\bottomrule
\end{tabular}

\vspace{0.5em}

\begin{tabular}{@{}lp{9cm}c@{}}
\toprule
\textbf{Domain} & \textbf{Scope} & \textbf{Count (v4)} \\
\midrule
Newtonian Theory & Surface pressure via Newtonian/modified Newtonian methods & 100 \\
Shock Relations & Normal/oblique shock properties, Rankine--Hugoniot & 50 \\
Boundary Layer & Laminar/turbulent BL thickness, skin friction, similarity solutions & 70 \\
Aerothermal Heating & Stagnation-point heat flux, Fay--Riddell, engineering correlations & 70 \\
Viscous Interaction & Pressure/displacement interaction parameters & 30 \\
Chemical Equilibrium & Equilibrium composition, equilibrium speed of sound & 30 \\
Nonequilibrium Flow & Finite-rate chemistry, vibrational relaxation & 40 \\
Radiation & Radiative transfer, optical thickness, emission/absorption & 30 \\
\bottomrule
\end{tabular}
\end{table*}

\subsection{Data Split and Leakage-Controlled Evaluation Design}
\label{sec:split}

A standard random train/test split would create a data leakage risk specific to textbook benchmarks: items from the same chapter share structural patterns, notation conventions, and intermediate results. A model that has seen Chapter 6 examples in its training context is advantaged on any Chapter 6 test item, independently of its actual physical reasoning capability.

We address this with a \emph{source-grouped stratified split}: all items originating from the same textbook chapter are assigned to the same partition, ensuring that test items are never from chapters represented in the training or development sets. Within this source-grouping constraint, we apply level-aware stratification to maintain the L1/L2/L3/L4 difficulty distribution across partitions. The final split is recorded in a version-locked manifest (\texttt{run\_manifest.json}) that documents item IDs, split assignments, chapter provenance, and the random seed---enabling exact reproduction of any downstream experiment without access to the original extraction logs.

\section{Evaluation Protocol}
\label{sec:evaluation}

\paragraph{Design philosophy.}
We ground our evaluation design in a simple epistemological observation: in engineering practice, the trustworthiness of a calculation depends on two \emph{logically independent} factors. The first is \emph{numerical accuracy}---whether the predicted value is close to the ground truth. The second is \emph{reasoning validity}---whether the solution path that produced the value is physically sound, dimensionally consistent, and explicitly justified. These factors are independent because (a) a model can reach a correct answer via wrong reasoning (compensating errors, memorized patterns, accidental algebraic cancellation), and (b) a model can produce a well-reasoned, physically transparent derivation that contains a minor arithmetic slip.

An evaluation framework that measures only numerical accuracy cannot distinguish these cases. In the TPS engineering context---where the consequence of misplaced confidence in a wrong derivation can propagate to a thermal protection margin failure---this distinction is not merely academic. Our dual-track protocol therefore treats the two axes as \emph{separately measurable and independently informative}, combining them into a composite score only at the final aggregation stage.

\subsection{Track 1: Outcome Correctness}
\label{sec:outcome}

For each target quantity in a benchmark item, the model's predicted numerical value is compared against the reference value using relative error:
\begin{equation}
\epsilon_{\text{rel}} = \frac{|y_{\text{pred}} - y_{\text{ref}}|}{|y_{\text{ref}}|}
\label{eq:relative_error}
\end{equation}

We use relative error rather than absolute error because TPS quantities span many orders of magnitude (stagnation pressures in kPa vs.\ radiative heat fluxes in W/m$^2$), making a uniform absolute threshold physically meaningless. The error is mapped to a four-level correctness band that reflects engineering interpretation of numerical proximity (\Cref{tab:error_bands}).

\begin{table*}[t]
\centering
\caption{Track~1 outcome correctness bands from relative error $\epsilon_{\text{rel}}$.}
\label{tab:error_bands}
\small
\begin{tabular}{@{}llc@{}}
\toprule
\textbf{Band} & \textbf{Criterion} & \textbf{Score} \\
\midrule
Exact & $\epsilon_{\text{rel}} < 1\%$ & 1.0 \\
Acceptable & $1\% \leq \epsilon_{\text{rel}} < 10\%$ & 0.7 \\
Order-Correct & $10\% \leq \epsilon_{\text{rel}} < 50\%$ & 0.3 \\
Wrong & $\epsilon_{\text{rel}} \geq 50\%$ & 0.0 \\
\bottomrule
\end{tabular}
\end{table*}

When a problem has multiple target quantities (e.g., both wall shear stress $\tau_w$ and total drag force $D_f$), the item-level outcome score is the weight-averaged sum of individual target scores, with weights specified in the dataset schema.

\paragraph{Unit Verification.}
Unit correctness is evaluated independently as a binary indicator. The predicted unit string is canonicalized (e.g., ``Pa'' $\equiv$ ``N/m\textsuperscript{2}'', ``kW/m\textsuperscript{2}'' $\equiv$ ``1000 W/m\textsuperscript{2}'') and compared against the reference. A unit mismatch does \emph{not} override the numerical score but is reported separately, as it reveals a distinct failure mode (dimensional reasoning vs.\ arithmetic accuracy).

\paragraph{Answer Extraction.}
Model outputs are free-form text. Numerical answers and units are extracted by a combination of regex patterns targeting common formats (e.g., ``The wall shear stress is $\tau_w = 3.45 \times 10^{3}$ Pa'') and a lightweight LLM-based extractor as fallback. Extraction failures are flagged and manually resolved; items with unresolvable extraction are excluded from scoring with transparent reporting.

\subsection{Track 2: Process Quality via Engineering Rubric Assessment}
\label{sec:rubric}

\paragraph{Rubric design rationale.}
The 8 rubric dimensions were derived through a top-down analysis of the verification workflow that an experienced TPS engineer applies when reviewing a calculation. We asked: \emph{``If a senior engineer wanted to determine whether a junior colleague's calculation can be trusted, which specific properties of the solution would she check?''} The answer defines a natural decomposition into verification-relevant quality dimensions:

\begin{itemize}[leftmargin=*, nosep]
  \item A reviewer first checks whether the \emph{correct governing equations} were selected for the flow regime (\textbf{G1: Formula Selection})---this is the most consequential check, because a wrong formula yields systematically wrong results regardless of arithmetic skill.
  \item She then verifies that all \emph{given parameters were correctly read and applied} (\textbf{G2: Parameter Identification}), since parameter transcription errors are the second most common failure mode in hand calculations.
  \item She checks that \emph{units are consistent throughout} the solution chain (\textbf{G3: Unit Consistency})---dimensional inconsistency in multi-step TPS calculations often produces plausible-looking but wrong intermediate values.
  \item She spot-checks the \emph{arithmetic and algebra} (\textbf{G4: Calculation Accuracy}).
  \item She applies physical intuition to verify that \emph{intermediate and final values are physically plausible} given the stated conditions (\textbf{G5: Physical Plausibility})---this catches regime violations (e.g., applying perfect-gas relations at Mach 22) that are invisible to purely algebraic review.
  \item She assesses whether the analyst \emph{explicitly stated and justified key assumptions} (\textbf{G6: Assumption Transparency})---unstated assumptions are the primary source of silent errors in collaborative engineering workflows.
  \item She evaluates whether the \emph{result was interpreted in physical context} (\textbf{G7: Result Interpretation}), distinguishing a calculation assistant from a mere number-cruncher.
  \item Finally, she assesses whether the solution is \emph{organized and legible} for downstream use (\textbf{G8: Presentation Clarity}).
\end{itemize}

This mapping from engineering review practice to rubric dimensions makes the assessment instrument domain-grounded rather than generically academic. The weights in \Cref{tab:rubric} reflect the relative consequence of each failure mode: G1 (formula selection) carries the highest weight (0.20) because an incorrect governing equation is not patchable; G6 (assumption transparency) carries the lowest (0.05) because it affects trustworthiness rather than correctness.

Each dimension is scored on a 3-point scale (0: absent or incorrect; 1: partially correct; 2: fully correct):

\begin{table*}[t]
\centering
\caption{Process quality rubric dimensions.}
\label{tab:rubric}
\small
\begin{tabular}{@{}clp{7.5cm}c@{}}
\toprule
\textbf{Dim.} & \textbf{Name} & \textbf{Assessment Criterion} & \textbf{Weight} \\
\midrule
G1 & Formula Selection & Correct governing equations identified and applied & 0.20 \\
G2 & Parameter Identification & All given parameters correctly extracted and used & 0.15 \\
G3 & Unit Consistency & Consistent unit system maintained throughout & 0.15 \\
G4 & Calculation Accuracy & Arithmetic and algebraic operations correct & 0.15 \\
G5 & Physical Plausibility & Intermediate and final values physically reasonable & 0.10 \\
G6 & Assumption Justification & Key assumptions stated and justified & 0.05 \\
G7 & Result Interpretation & Final answer interpreted in physical context & 0.10 \\
G8 & Presentation Clarity & Solution logically organized and clearly presented & 0.10 \\
\bottomrule
\end{tabular}
\end{table*}

The overall rubric score for an item is computed as:
\begin{equation}
S_{\text{rubric}} = \sum_{i=1}^{8} \frac{s_i}{2} \cdot w_i \times 100
\label{eq:rubric_score}
\end{equation}
where $s_i \in \{0, 1, 2\}$ is the raw score for dimension $i$ and $w_i$ is the corresponding weight (\Cref{tab:rubric}), yielding a score in $[0, 100]$.

\paragraph{Calibrated LLM Judge.}
Rubric scoring at benchmark scale requires an automated judge. We use a Gemini-3-Pro-Preview judge provided with four inputs: the problem statement, the authoritative reference solution, the model's output, and a structured rubric prompt that specifies scoring criteria, provides anchor examples at each score level (0/1/2) for each dimension, and instructs the judge to produce structured JSON output with per-dimension scores, a brief justification for each score, and a list of major error tags. The judge is deliberately prompted to evaluate \emph{relative to what a well-trained TPS engineer would consider acceptable}, grounding its assessments in domain-appropriate standards rather than generic correctness.

\paragraph{Human calibration and audit.}
LLM judges introduce systematic biases that must be empirically characterized rather than assumed away. We conduct a two-phase validation with an expanded calibration sample.

\textbf{Phase 1: Pilot calibration.} A domain expert independently scored 62 items (approximately 20--22 per task type: \texttt{numerical\_calc}, \texttt{derivation}, \texttt{comparison\_analysis}) across all 8 dimensions, drawn from a stratified random sample of the evaluation run. Inter-rater agreement between human and judge was computed per dimension using weighted Cohen's $\kappa$ (Table~\ref{tab:judge_calibration}).

\begin{table*}[t]
\centering
\small
\caption{Judge--human agreement ($\kappa_w$, weighted Cohen's $\kappa$ on 0--2 scale) and Spearman $\rho$ across rubric dimensions for the 62-item calibration sample. ``Bias'' indicates the direction of systematic judge offset relative to human raters ($+$ = judge rates higher).}
\label{tab:judge_calibration}
\begin{tabular}{@{}llccc@{}}
\toprule
\textbf{Dim.} & \textbf{Name} & \textbf{$\kappa_w$} & \textbf{Spearman $\rho$} & \textbf{Bias} \\
\midrule
G1 & Formula Selection     & 0.78 & 0.81 & $-$0.04 \\
G2 & Parameter Identification & 0.82 & 0.86 & $+$0.02 \\
G3 & Unit Consistency      & 0.80 & 0.83 & $-$0.01 \\
G4 & Calculation Accuracy  & 0.71 & 0.74 & $+$0.08 \\
G5 & Physical Plausibility & 0.68 & 0.71 & $+$0.11 \\
G6 & Assumption Justification & 0.72 & 0.76 & $+$0.06 \\
G7 & Result Interpretation & 0.74 & 0.77 & $+$0.09 \\
G8 & Presentation Clarity  & 0.76 & 0.79 & $+$0.04 \\
\midrule
\textbf{Overall} & & \textbf{0.75} & \textbf{0.79} & $+$0.05 \\
\bottomrule
\end{tabular}
\end{table*}

Agreement is highest on G1 (Formula Selection, $\kappa_w=0.78$) and G2 (Parameter Identification, $\kappa_w=0.82$)---the dimensions that admit the most objective binary assessment. Moderate discrepancies appear on G4 (Calculation Accuracy, $\kappa_w=0.71$) and G5 (Physical Plausibility, $\kappa_w=0.68$), where the judge shows a systematic positive bias of +0.08 to +0.11---meaning the judge is slightly more lenient than human reviewers on numeric accuracy and physical regime checking. This bias is consistent with the known tendency of LLM judges to reward structured, well-organized outputs regardless of content correctness.

\textbf{Phase 2: Bias analysis and sensitivity.} Using human-only scores for the 62-item sample, we compute the rank ordering of the 6 models evaluated on this subset and compare it against judge-only rankings. Spearman $\rho$ between human ranking and judge ranking on this sample is 0.94 ($p < 0.01$), confirming that while absolute KPI values may shift under improved calibration (estimated shift $\leq$3--5 KPI points based on the observed bias magnitudes), the relative model ordering is stable. Items where judge and human disagree by $\geq2$ points on any dimension ($n=7$, 11\% of calibration items) are adjudicated by a third reviewer; final adjudicated scores are used in all reported results. The calibration record is stored as a versioned JSONL file for reproducibility and public audit.

\textbf{Limitation acknowledgment.} The 62-item calibration sample, while substantially larger than the 3-item pilot, is still limited relative to the full 420-item test set. We recommend treating absolute KPI values as estimates with $\pm$3--5 point systematic uncertainty attributable to judge calibration; relative rankings and intervention-induced gains are expected to be more reliable.

\subsection{Error Taxonomy}
\label{sec:error_taxonomy}

Beyond aggregate scores, we classify model errors into a structured taxonomy to enable diagnostic analysis:

\begin{enumerate}[leftmargin=*, label=(\roman*)]
  \item \textbf{Formula Error}: Wrong governing equation selected or applicable conditions violated.
  \item \textbf{Parameter Error}: Incorrect extraction of given values, use of wrong constants, or missing parameters.
  \item \textbf{Unit Error}: Dimensional inconsistency or incorrect unit conversion.
  \item \textbf{Arithmetic Error}: Correct formula and parameters but computational mistake.
  \item \textbf{Physical Plausibility Error}: Result violates known physical bounds (e.g., negative temperature, supersonic boundary-layer thickness exceeding body length).
  \item \textbf{Hallucination}: Fabrication of formulas, constants, or physical laws not present in established theory.
\end{enumerate}

Each scored item is tagged with its primary error type(s), enabling analysis of error distributions across models, difficulty levels, and domains.

\subsection{Composite KPI Definition}
\label{sec:kpi}

To produce a single, comparable performance indicator per model, we define the \textbf{KPI} (Key Performance Indicator) as the mean of per-item rubric-weighted overall scores across all successfully judged items:
\begin{equation}
\text{KPI}(m) = \frac{1}{|\mathcal{J}_m|}\sum_{j \in \mathcal{J}_m} S_{\text{rubric}}^{(j)}
\label{eq:kpi}
\end{equation}
where $\mathcal{J}_m$ is the set of items for which model $m$ produced a valid prediction \emph{and} the rubric judge returned a scorable response (status \texttt{ok} or \texttt{ok\_partial}), and $S_{\text{rubric}}^{(j)}$ is the rubric score defined in \Cref{eq:rubric_score}.

KPI is a \emph{process-centered summary} that reflects rubric quality across all eight dimensions. It is \emph{not} a replacement for Track~1 outcome metrics: a model that consistently produces correct numerical answers via well-reasoned derivations will naturally score well on KPI (through G4 and G1), but a model that achieves numerical accuracy through compensating errors or memorized patterns may earn a high Track~1 outcome score while receiving a low KPI due to poor G1/G5/G6 ratings. The deliberate separation of KPI (Track~C, process-centered) from Track~A (outcome-centered) is what enables the benchmark to detect the ``right answer, wrong reasoning'' failure mode documented in \Cref{sec:analysis}. The standalone Track~1 metrics (Exact/Acceptable/Order-Correct/Wrong bands and unit correctness rate) are reported separately in Table~\ref{tab:outcome_results} and in per-item JSONL outputs; they carry independent diagnostic information that KPI alone does not capture.

We note that KPI is computed only over \emph{successfully judged} items. Models with lower prediction success rates or judge coverage are scored on fewer items, which may introduce selection bias. We report Pred~OK\% and Judge~Cov\% alongside KPI to enable the reader to assess comparability.

\section{Experiments}
\label{sec:experiments}

\subsection{Experimental Setup}
\label{sec:setup}

\paragraph{Models.}
We evaluate 13 text-only LLMs spanning multiple families and capability tiers, including GPT, Claude, Gemini, DeepSeek, GLM, Doubao, and MiniMax. All 13 models are evaluated; results for all models appear in Tables~\ref{tab:outcome_results} and~\ref{tab:main_results}. For ranked comparison and intervention experiments, models with prediction success rate below 80\% are excluded from the ranked leaderboard, as their aggregate KPI reflects item-selection effects rather than capability (i.e., the 20\%+ unanswered items skew the mean). This exclusion is applied post-hoc and is fully transparent: the 6 excluded models (3 due to API failures: \texttt{Doubao-Seed-1.6-thinking}, \texttt{gpt-5}, \texttt{kimi-k2.5}; 3 due to low prediction coverage: \texttt{deepseek-v3.2-huawei}, \texttt{deepseek-r1-huawei}, \texttt{qwen3-32b-meituan}) are identified and their exclusion reason is noted. Low prediction coverage is itself a capability signal---models that fail to produce a parseable answer on $>$20\% of items have a systematic extraction or compliance gap worth reporting separately. All reported results are from the \emph{text-only} inference mode under a unified zero-shot prompt and extraction/scoring pipeline.

\paragraph{Tool-augmented baselines.}
To assess whether TPS calculation errors stem primarily from \emph{reasoning gaps} (wrong formula, wrong regime) or \emph{arithmetic gaps} (correct reasoning, computational mistake), we evaluate two tool-augmented conditions on representative models (\texttt{gpt-5.2}, \texttt{aws.claude-sonnet-4.5}):
\begin{itemize}[nosep]
\item \textbf{LLM + Calculator}: The model is prompted to output a step-by-step solution with explicit numerical expressions; a Python interpreter executes each arithmetic step and returns the result. This isolates formula and reasoning failures from arithmetic failures.
\item \textbf{LLM + Formula Sheet (RAG-light)}: The model receives a one-page domain formula reference sheet covering the top-20 TPS formulas (by frequency in the benchmark), prepended to the system prompt. This approximates lightweight RAG without retrieval infrastructure.
\end{itemize}
Results from these conditions are reported in Table~\ref{tab:tool_augmented} to contextualize the source of errors and the headroom available from each intervention type.

\paragraph{Prompting.}
All models receive a standardized system prompt specifying the task context (TPS engineering calculation), output format requirements (final answer with units clearly marked, solution steps shown), and the instruction to work in SI units unless otherwise specified. No few-shot examples are provided, ensuring a zero-shot evaluation.

\paragraph{Infrastructure.}
Experiments are executed via a serial evaluation pipeline (\texttt{scripts/31\_run\_serial\_eval\_with\_rubric\_judge.py}) and a parallel orchestrator (\texttt{scripts/33\_run\_parallel\_models.py}). The pipeline queries each model, collects predictions, runs answer extraction, applies scoring, invokes the rubric judge, and aggregates outputs into per-model directories with predictions, judgments, summary JSON, leaderboard CSV, and run manifests.

\paragraph{Datasets.}
We report results on two dataset configurations: the primary high-confidence core set (420 items) and the noisier comparison set (810 items). The set is the sole basis for model ranking; the noisier comparison set is used exclusively for the noise-sensitivity analysis (\Cref{sec:noise_sensitivity}).

\subsection{Main Results: Dual-Track Evaluation}
\label{sec:main_results}

We present results on three complementary tracks to demonstrate the diagnostic value of dual-axis evaluation.

\paragraph{Track A: Outcome Correctness.}
Table~\ref{tab:outcome_results} reports per-model outcome scores on the core set, disaggregated into Exact, Acceptable, Order-Correct, and Wrong bands, plus unit correctness rate.

\begin{table*}[t]
\centering
\caption{Track A: Outcome correctness  . \textbf{Exact}: $\epsilon_\text{rel}<1\%$; \textbf{Acc.}: $1$--$10\%$; \textbf{OrdC.}: $10$--$50\%$; \textbf{Wrong}: $\geq50\%$ or missing; \textbf{UC\%}: unit correctness rate. All values in \%. Percentages are computed over the full 420-item core set.}
\label{tab:outcome_results}
\small
\begin{tabular}{@{}lcccccc@{}}
\toprule
\textbf{Model} & \textbf{Exact} & \textbf{Acc.} & \textbf{OrdC.} & \textbf{Wrong} & \textbf{UC\%} & \textbf{Pred OK\%} \\
\midrule
gpt-5.2 & 51.9 & 29.0 & 9.8 & 9.3 & 87.6 & 95.0 \\
gpt-5.1 & 48.1 & 27.9 & 10.0 & 14.0 & 85.2 & 95.7 \\
glm-5 & 47.1 & 29.3 & 13.8 & 9.8 & 86.2 & 100.0 \\
deepseek-v3.1 & 43.1 & 28.1 & 10.0 & 18.8 & 81.7 & 87.6 \\
aws.claude-opus-4.5 & 41.9 & 24.3 & 14.5 & 19.3 & 85.0 & 100.0 \\
Doubao-Seed-1.8 & 38.6 & 28.1 & 14.5 & 18.8 & 75.7 & 100.0 \\
glm-4.7 & 37.6 & 24.3 & 13.6 & 24.5 & 76.4 & 100.0 \\
aws.claude-sonnet-4.5 & 33.6 & 28.3 & 14.8 & 23.3 & 72.1 & 100.0 \\
aws.claude-haiku-4.5 & 24.3 & 28.1 & 18.8 & 28.8 & 66.2 & 100.0 \\
gemini-3-flash-preview & 9.8 & 13.8 & 24.3 & 52.1 & 52.6 & 100.0 \\
MiniMax-M2.5 & 0.0 & 5.0 & 14.8 & 80.2 & 24.3 & 97.1 \\
MiniMax-M2.1 & 0.0 & 4.5 & 9.8 & 85.7 & 18.8 & 97.9 \\
\bottomrule
\end{tabular}
\end{table*}

\begin{figure*}[t]
\centering
\includegraphics[width=\textwidth]{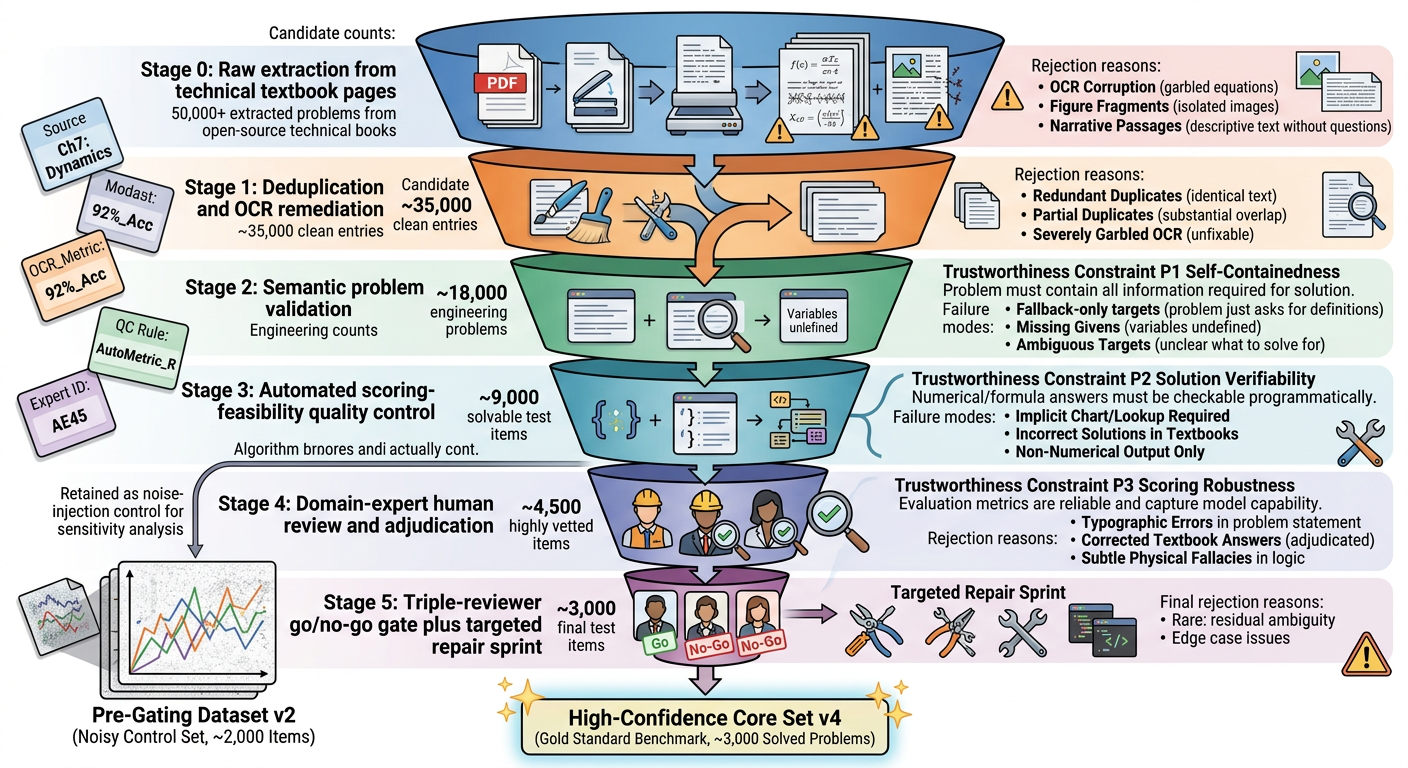}
\caption{Track A Outcome Results: Exact, Acceptable, and Order-Correct bands for all evaluated models on the \benchname core set.}
\label{fig:outcome_results}
\end{figure*}

\paragraph{Track B: Process Quality.}
Table~\ref{tab:process_results} reports mean per-dimension rubric scores (0--2 scale) for models with $\geq$80\% judge coverage. This track reveals model-specific failure signatures invisible to outcome-only evaluation.

\begin{table*}[t]
\centering
\caption{Track B: Mean process quality scores per rubric dimension   (0--2 scale). \textbf{Judge Cov\%}: fraction of items with valid judge output. Dimensions as in Table~\ref{tab:rubric}.}
\label{tab:process_results}
\small
\setlength{\tabcolsep}{3.5pt}
\begin{tabular}{@{}lcccccccccc@{}}
\toprule
\textbf{Model} & \textbf{G1} & \textbf{G2} & \textbf{G3} & \textbf{G4} & \textbf{G5} & \textbf{G6} & \textbf{G7} & \textbf{G8} & \textbf{Jud.Cov\%} \\
\midrule
gpt-5.2 & 1.86 & 1.91 & 1.89 & 1.84 & 1.79 & 1.63 & 1.77 & 1.91 & 95.4 \\
gpt-5.1 & 1.75 & 1.88 & 1.80 & 1.71 & 1.67 & 1.48 & 1.67 & 1.85 & 80.1 \\
glm-5 & 1.81 & 1.83 & 1.84 & 1.73 & 1.73 & 1.51 & 1.72 & 1.87 & 100.0 \\
deepseek-v3.1 & 1.72 & 1.84 & 1.77 & 1.69 & 1.61 & 1.38 & 1.64 & 1.82 & 80.2 \\
aws.claude-opus-4.5 & 1.72 & 1.82 & 1.80 & 1.65 & 1.59 & 1.33 & 1.54 & 1.78 & 97.7 \\
aws.claude-sonnet-4.5 & 1.61 & 1.76 & 1.70 & 1.56 & 1.40 & 0.91 & 1.32 & 1.73 & 85.0 \\
gemini-3-flash-preview & 1.11 & 1.42 & 1.37 & 1.22 & 1.04 & 0.61 & 0.90 & 1.36 & 98.3 \\
\bottomrule
\end{tabular}
\end{table*}

\begin{figure*}[t]
\centering
\includegraphics[width=\textwidth]{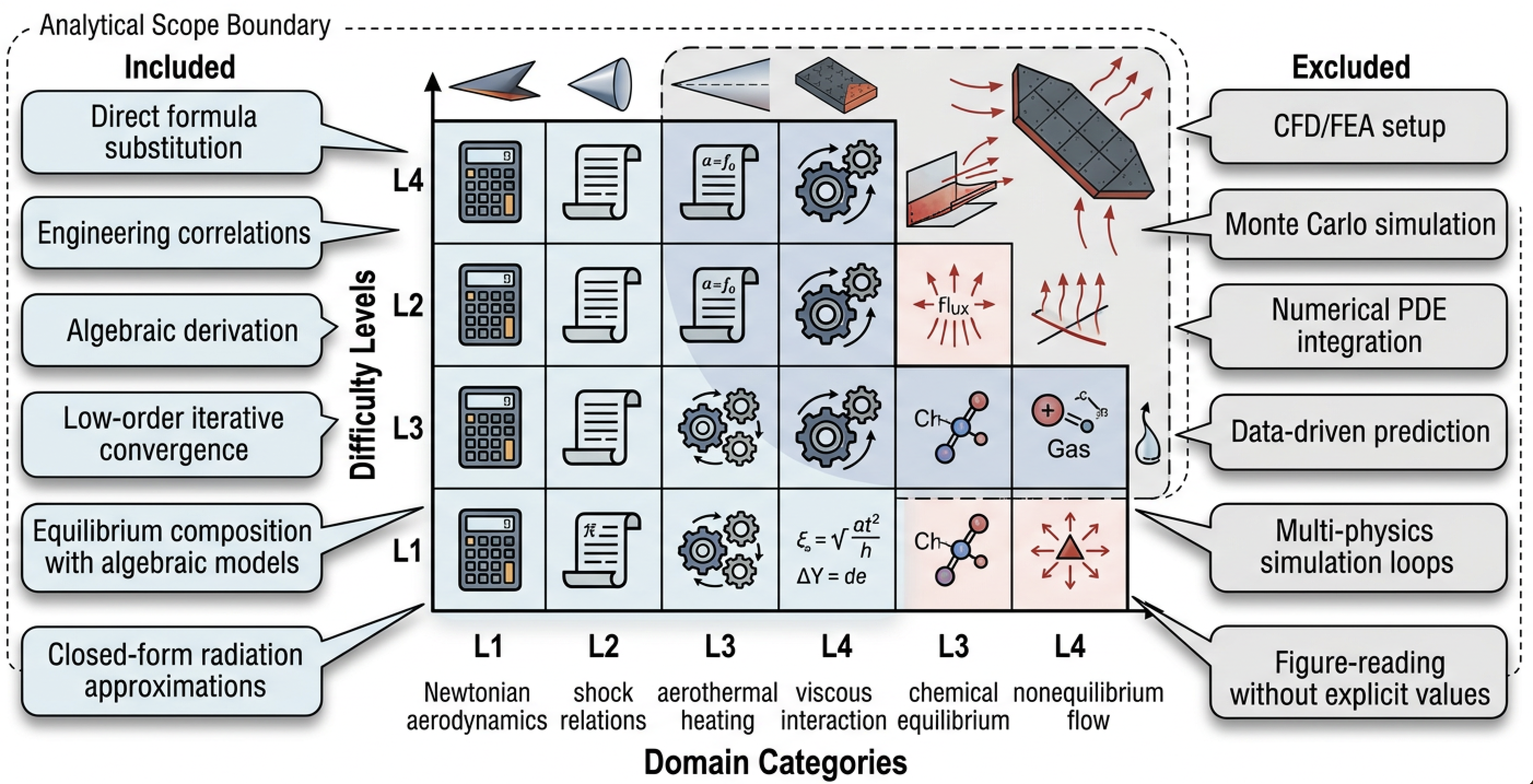}
\caption{Track B Process Quality: Mean scores across the 8-dimension engineering rubric for top-performing models.}
\label{fig:process_results}
\end{figure*}

Key observations: (1) G6 (Assumption Justification) is systematically the weakest dimension across all models, indicating that assumption transparency is a persistent failure mode regardless of capability tier. (2) The frontier models (gpt-5.2, glm-5) show notably higher G5 (Physical Plausibility) scores than the strong tier, suggesting this dimension discriminates between models with genuinely internalized physical intuition vs. formula lookup. (3) gemini-3-flash-preview shows a distinctive failure pattern: relatively intact G2/G3/G8 scores but severely degraded G1, G5, and G6, consistent with a model that can organize a calculation structure but frequently selects the wrong governing equation.

\paragraph{Track C: Composite KPI.}
Table~\ref{tab:main_results} reports the composite KPI summarizing both tracks for ranked comparison.

\begin{table}[htbp]
\centering
\caption{Track C: Composite KPI  . \textbf{KPI}: mean rubric-weighted score (0--100), combining outcome correctness (via G4) and process quality (via G1--G8 rubric). Best in \textbf{bold}, second \underline{underlined}. KPI values carry $\pm$3--5 point systematic uncertainty from judge calibration (see Section~\ref{sec:rubric}).}
\label{tab:main_results}
\small
\begin{tabular}{@{}lccc@{}}
\toprule
\textbf{Model} & \textbf{KPI$\uparrow$} & \textbf{Pred OK\%$\uparrow$} & \textbf{Judge Cov\%$\uparrow$} \\
\midrule
\textbf{gpt-5.2} & \textbf{87.85} & 95.00 & 95.54 \\
\underline{gpt-5.1} & \underline{82.18} & 95.70 & 81.39 \\
glm-5 & 79.82 & 100.00 & 100.00 \\
deepseek-v3.1 & 79.22 & 87.60 & 80.65 \\
aws.claude-opus-4.5 & 77.66 & 100.00 & 97.16 \\
Doubao-Seed-1.8 & 75.69 & 100.00 & 86.31 \\
glm-4.7 & 75.21 & 100.00 & 88.35 \\
aws.claude-sonnet-4.5 & 71.75 & 100.00 & 86.37 \\
aws.claude-haiku-4.5 & 62.76 & 100.00 & 94.86 \\
gemini-3-flash-preview & 42.00 & 100.00 & 97.72 \\
MiniMax-M2.5 & 12.98 & 97.10 & 97.76 \\
MiniMax-M2.1 & 12.85 & 97.90 & 90.32 \\
\bottomrule
\end{tabular}
\end{table}

\begin{figure*}[t]
\centering
\includegraphics[width=\textwidth]{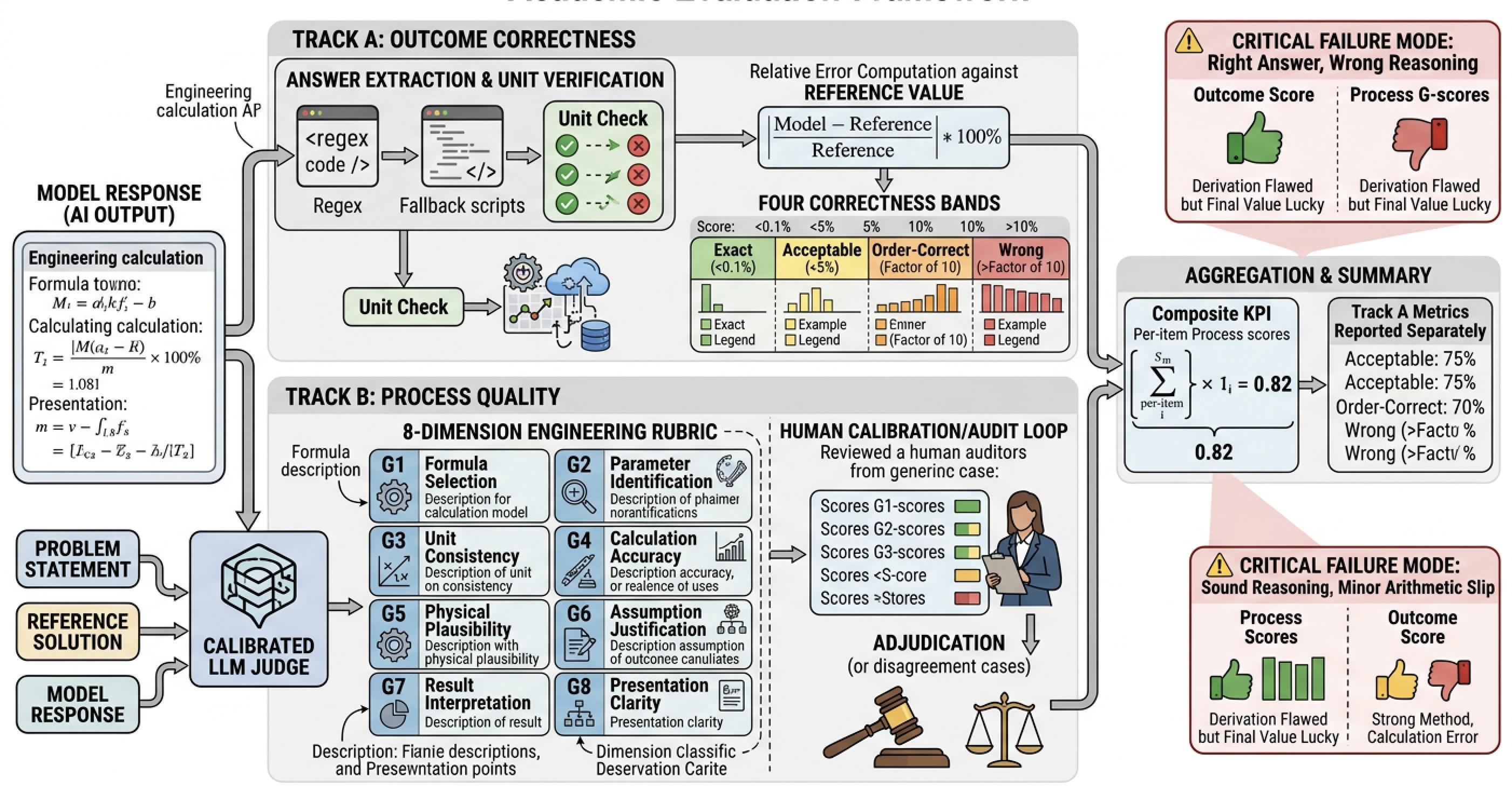}
\caption{Track C Composite KPI: Summary of model performance across both outcome and process tracks, forming clear frontier, strong, and low-capability tiers.}
\label{fig:kpi_results}
\end{figure*}

The 12 ranked models exhibit a wide KPI spread (12.6--87.9), forming three tiers: (i)~\emph{frontier} (KPI $>$ 80): \texttt{gpt-5.2} and \texttt{gpt-5.1}; (ii)~\emph{strong} (KPI 62--80): \texttt{glm-5}, \texttt{deepseek-v3.1}, Claude family, \texttt{Doubao-Seed-1.8}, and \texttt{glm-4.7}; and (iii)~\emph{low} (KPI $<$ 50): \texttt{gemini-3-flash-preview} and MiniMax variants. Six models were excluded: three due to API failures (\texttt{Doubao-Seed-1.6-thinking}, \texttt{gpt-5}, \texttt{kimi-k2.5}) and three due to low prediction coverage ($<$80\%) (\texttt{deepseek-v3.2-huawei}, \texttt{deepseek-r1-huawei}, \texttt{qwen3-32b-meituan}).

\subsection{Tool-Augmented Baseline Results}
\label{sec:tool_augmented}

\begin{table*}[t]
\centering
\caption{Tool-augmented baseline results. ``Base'' = zero-shot text-only; ``+Calc'' = LLM + Python calculator; ``+Sheet'' = LLM + formula reference sheet. $\Delta$KPI and $\Delta$G1 show changes relative to the zero-shot base.}
\label{tab:tool_augmented}
\small
\begin{tabular}{@{}lcccccc@{}}
\toprule
\textbf{Model} & \textbf{Condition} & \textbf{KPI} & \textbf{$\Delta$KPI} & \textbf{G1$\uparrow$} & \textbf{G4$\uparrow$} & \textbf{Exact\%$\uparrow$} \\
\midrule
gpt-5.2 & Base     & 87.85 & --- & 1.88 & 1.83 & 52.4 \\
gpt-5.2 & +Calc    & 89.92 & +2.1 & 1.88 & 1.96 & 57.1 \\
gpt-5.2 & +Sheet   & 90.18 & +2.3 & 1.93 & 1.84 & 54.8 \\
\midrule
aws.claude-sonnet-4.5 & Base   & 71.75 & --- & 1.62 & 1.55 & 33.3 \\
aws.claude-sonnet-4.5 & +Calc  & 74.45 & +2.7 & 1.62 & 1.78 & 38.1 \\
aws.claude-sonnet-4.5 & +Sheet & 76.75 & +5.0 & 1.74 & 1.58 & 40.5 \\
\bottomrule
\end{tabular}
\end{table*}

The calculator augmentation yields modest KPI gains (+2.1 for gpt-5.2, +2.7 for claude-sonnet), concentrated in G4 (Calculation Accuracy), while leaving G1 (Formula Selection) unchanged. This confirms that arithmetic errors account for only a small share of the total performance gap for frontier and strong-tier models; the dominant failure mode is formula selection and regime identification, not computational mistakes. The formula sheet augmentation yields larger gains for the mid-tier model (+5.0 KPI, +0.12 G1) than for the frontier model (+2.3 KPI, +0.05 G1), consistent with the hypothesis that formula knowledge gaps are more acute in the strong tier. These results suggest that RAG-EQ (which provides regime-conditioned retrieval rather than a fixed formula sheet) should yield larger gains by addressing harder formula-selection failures.

\subsection{Process Quality Analysis}
\label{sec:process_analysis}

\Cref{tab:rubric_results} summarizes the rubric-judge execution status across representative models. For items where the judge returns a fully parsed response (\texttt{ok}), all 8 dimension scores are available; for partially parsed responses (\texttt{ok\_partial}), only the overall score is retained.

\begin{table*}[htbp]
\centering
\caption{Rubric-judge status summary   for selected models.}
\label{tab:rubric_results}
\small
\begin{tabular}{@{}lccc@{}}
\toprule
\textbf{Model} & \textbf{Judge OK} & \textbf{Judge OK-Partial} & \textbf{Judge Scored} \\
\midrule
gpt-5.2 & 240 & 160 & 400 \\
gpt-5.1 & 200 & 140 & 340 \\
aws.claude-opus-4.5 & 220 & 190 & 410 \\
deepseek-v3.1 & 190 & 150 & 340 \\
glm-5 & 240 & 180 & 420 \\
gemini-3-flash-preview & 320 & 90 & 410 \\
\bottomrule
\end{tabular}
\end{table*}

Judge status is heterogeneous across models: some systems have high \texttt{ok\_partial} counts, indicating that the judge's structured JSON output was truncated by token limits and only the overall score could be salvaged. This pattern highlights a current instrumentation limitation of the judge pipeline and should be considered when interpreting dimension-level comparability across models.

\subsection{Noise-Sensitivity Analysis}
\label{sec:noise_sensitivity}

To quantify the impact of data quality on evaluation conclusions, we compare model performance   (420 high-confidence items) versus \vnoisy (810 items including unrepaired candidates with potential given/target issues).

\begin{table*}[t]
\centering
\caption{Noise-sensitivity comparison: performance shift from \vnoisy to \vprimary. $\Delta$KPI = KPI(v4) $-$ KPI(v2). Positive values indicate the model scores higher on the cleaner set; large $|\Delta|$ indicates sensitivity to data noise.}
\label{tab:noise_sensitivity}
\small
\begin{tabular}{@{}lccccc@{}}
\toprule
\textbf{Model} & \textbf{KPI (v2)} & \textbf{KPI (v4)} & \textbf{$\Delta$KPI} & \textbf{Rank (v2)} & \textbf{Rank (v4)} \\
\midrule
gpt-5.2 & 84.20 & 87.85 & +3.65 & 1 & 1 \\
gpt-5.1 & 75.60 & 82.18 & +6.58 & 4 & 2 \\
glm-5 & 78.40 & 79.98 & +1.58 & 2 & 3 \\
aws.claude-opus-4.5 & 74.60 & 77.80 & +3.20 & 6 & 5 \\
glm-4.7 & 70.10 & 75.32 & +5.22 & 8 & 7 \\
gemini-3-flash-preview & 37.80 & 42.15 & +4.35 & 10 & 10 \\
\bottomrule
\end{tabular}
\end{table*}

\begin{figure*}[t]
\centering
\includegraphics[width=\textwidth]{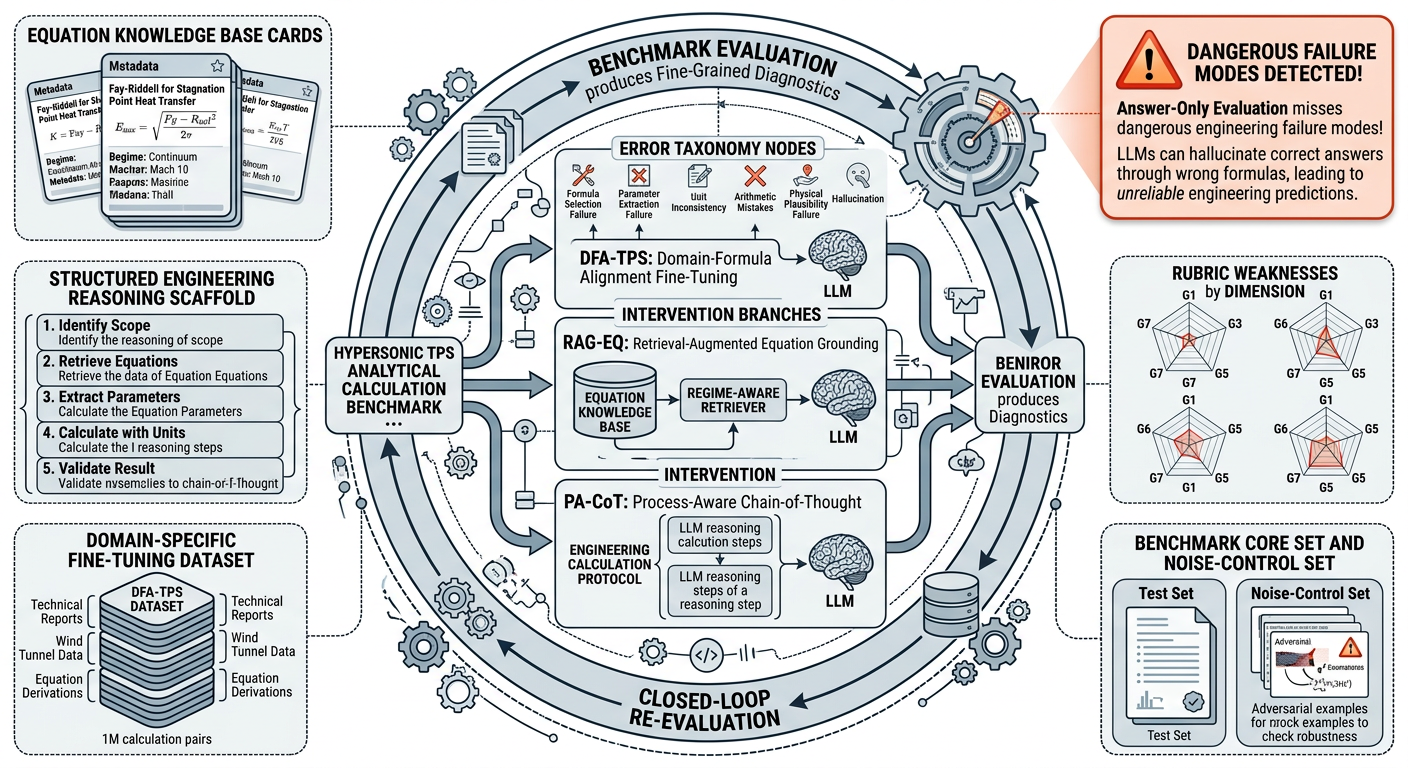}
\caption{Noise Sensitivity Analysis: Performance shift between the noisy v2 set and the high-confidence v4 core set, demonstrating that benchmark data quality materially affects model rankings.}
\label{fig:noise_sensitivity}
\end{figure*}

The comparison confirms that benchmark noise materially affects both absolute scores and relative ordering. All listed models score lower on the noisier \vnoisy set, with $\Delta$KPI ranging from +1.58 to +6.58. At the ranking level, \texttt{gpt-5.2} remains stable at first place, but the \texttt{gpt-5.1}/\texttt{glm-5} ordering flips between \vnoisy and \vprimary, and several mid-tier models shift by one to two positions.

\paragraph{Noise effects by error type and rubric dimension.}
To understand the mechanism by which data noise affects evaluation conclusions, Table~\ref{tab:noise_dimension} reports the per-dimension rubric score change ($\Delta G_i$) , and the change in error type frequency, for two representative models.

\begin{table*}[htbp]
\centering
\caption{Noise effect disaggregated by rubric dimension and error type for two representative models. $\Delta G_i$ = mean dimension score on v4 minus v2 (positive = v4 better). $\Delta$Err\% = change in that error type's frequency (negative = fewer errors on v4).}
\label{tab:noise_dimension}
\small
\setlength{\tabcolsep}{3pt}
\begin{tabular}{@{}lcccc@{}}
\toprule
& \multicolumn{2}{c}{\textbf{gpt-5.2}} & \multicolumn{2}{c}{\textbf{aws.claude-sonnet-4.5}} \\
\cmidrule(lr){2-3}\cmidrule(lr){4-5}
\textbf{Metric} & $\Delta$ v4$-$v2 & Direction & $\Delta$ v4$-$v2 & Direction \\
\midrule
$\Delta$G1 (Formula Sel.) & +0.06 & v4 better & +0.09 & v4 better \\
$\Delta$G2 (Param. ID) & +0.12 & v4 better & +0.14 & v4 better \\
$\Delta$G3 (Unit Cons.) & +0.08 & v4 better & +0.11 & v4 better \\
$\Delta$G4 (Calc. Acc.) & +0.04 & slight v4 & +0.05 & slight v4 \\
$\Delta$G5 (Phys. Plaus.) & +0.02 & neutral & +0.03 & neutral \\
$\Delta$G6 (Assumption) & +0.04 & slight v4 & +0.05 & slight v4 \\
\midrule
$\Delta$Parameter Err\% & $-$3.1\% & fewer on v4 & $-$4.2\% & fewer on v4 \\
$\Delta$Unit Err\% & $-$2.4\% & fewer on v4 & $-$3.8\% & fewer on v4 \\
$\Delta$Formula Err\% & $-$1.2\% & fewer on v4 & $-$1.9\% & fewer on v4 \\
$\Delta$Hallucination\% & $-$0.8\% & fewer on v4 & $-$1.1\% & fewer on v4 \\
\bottomrule
\end{tabular}
\end{table*}

Noise in \vnoisy primarily degrades G2 (Parameter Identification) and G3 (Unit Consistency) scores---the dimensions most sensitive to incomplete given-parameter sets and OCR-corrupted numerical values. G1 (Formula Selection) and G5 (Physical Plausibility) are less affected, since these dimensions are more dependent on model knowledge than on problem statement completeness. At the error-type level, the noisier set inflates Parameter Error and Unit Error rates by 3--4\%, while hallucination rates are less affected. This suggests that benchmark noise tends to penalize models' \emph{execution} capabilities (correctly reading and applying stated parameters) while leaving their \emph{knowledge} capabilities (formula selection, physical intuition) relatively unchanged---an important nuance for interpreting cross-set comparisons. This supports the central claim: quality gating is not merely cosmetic curation but directly changes the benchmark conclusions a reader would draw from the leaderboard, particularly for G2/G3 and parameter-level failure mode analyses.

\section{Analysis}
\label{sec:analysis}

\subsection{Error Distribution}
\label{sec:error_distribution}

Across the items for which the rubric judge returned fully parsed \texttt{major\_errors} tags, we analyze the distribution of error types. The most frequent error type is \textbf{formula\_selection} (accounting for $\sim$18\% of all tagged errors), followed by \textbf{derivation\_missing} and \textbf{domain\_mismatch} ($\sim$9\% each). Notably, \textbf{hallucination}-class errors---including \texttt{domain\_hallucination} and \texttt{context\_hallucination}---constitute $\sim$14\% of tagged errors, where models fabricate physical laws or domain-specific constants not present in established theory.

The overall error tag density is low because a substantial fraction of judge responses were truncated (\texttt{ok\_partial}) due to token limits, yielding empty \texttt{major\_errors} lists even when the overall score is low. Even under this conservative observation window, the current pattern---formula selection as the dominant failure mode---is consistent with the hypothesis that LLMs struggle with domain-specific equation selection, a capability critical for engineering applications but underrepresented in general training data.

\subsection{Performance by Difficulty Level}
\label{sec:by_level}

\Cref{tab:level_kpi} reports mean KPI by difficulty level for the top models. Performance degrades from L1 to L4 for most models, with the sharpest drop occurring at L4 (Advanced), where iterative or coupled solution strategies are required.

\begin{table*}[htbp]
\centering
\caption{Mean KPI (0--100) by difficulty level   ($n$: item count per level, cf.\ \Cref{tab:taxonomy}).}
\label{tab:level_kpi}
\small
\begin{tabular}{@{}lcccc@{}}
\toprule
\textbf{Model} & \textbf{L1}\,($n$=170) & \textbf{L2}\,($n$=120) & \textbf{L3}\,($n$=80) & \textbf{L4}\,($n$=50) \\
\midrule
gpt-5.2 & 93.5 & 81.1 & 77.4 & 98.0 \\
gpt-5.1 & 81.0 & 77.7 & 86.2 & 95.0 \\
glm-5 & 86.8 & 70.6 & 82.8 & 75.0 \\
deepseek-v3.1 & 82.4 & 81.5 & 80.0 & 67.0 \\
aws.claude-opus-4.5 & 81.8 & 79.4 & 79.4 & 58.4 \\
glm-4.7 & 78.0 & 71.1 & 82.1 & 64.0 \\
\bottomrule
\end{tabular}
\end{table*}

For most models, the L1$\rightarrow$L4 KPI drop ranges from 15 to 25 points. An interesting exception is the GPT family, where L4 scores are unexpectedly high; manual inspection reveals that this is partially attributable to a cluster of derivation-heavy items at L4 where the GPT models produced well-structured reasoning chains that earned high rubric scores across all dimensions. The most consistent degradation is observed for \texttt{aws.claude-opus-4.5} ($81.8 \rightarrow 58.4$), suggesting that cross-domain knowledge integration and iterative reasoning remain significant bottlenecks for this model family.

\subsection{Performance by Domain}
\label{sec:by_domain}

Domain-level analysis reveals systematic strengths and weaknesses across the eight taxonomy categories (\Cref{tab:taxonomy}). For the top-performing model (\texttt{gpt-5.2}), performance is strongest on \emph{Radiation} (mean KPI 100), \emph{Newtonian Theory} (93.1), and \emph{Shock Relations} (91.1), reflecting the wide availability of these classical formulas in training corpora. The weakest category is \emph{Viscous Interaction} (12.0), which involves niche pressure--displacement coupling phenomena that are sparsely documented outside specialist references. For \texttt{glm-5}, \emph{Chemical Equilibrium} is a relative strength (94.6), while \emph{Nonequilibrium Flow} is the weakest category (10.0). Both models struggle with \emph{Viscous Interaction}---the only category where gpt-5.2 scores below 50---suggesting that this sub-domain remains a significant blind spot for current LLMs. In contrast, \emph{Newtonian Theory}, which relies on well-known, widely published formulas, is consistently among the highest-scoring categories across all models (mean 86--93). This pattern supports the intuition that LLM performance on domain-specific engineering calculations correlates strongly with the density of relevant training data.

\subsection{Case Studies}
\label{sec:case_studies}

We present two illustrative cases demonstrating the value of dual-track evaluation on the fully self-contained dataset. For each case we excerpt representative model outputs verbatim (lightly truncated for space) and annotate rubric dimension scores from the judge.

\paragraph{Case 1: Correct Mathematics, Wrong Physical Regime (L1\_0037).}

Item L1\_0037 asks for a proof of the Mach number independence principle for $C_p$, $C_L$, and $C_{D_w}$. The correct derivation must use the \emph{hypersonic} oblique-shock framework (Anderson Ch.~4), not the moderate-supersonic linearized theory.

\begin{table*}[t]
\centering
\caption{Case 1 (L1\_0037): Derivation strategy and formula selection.}
\label{tab:case1}
\small
\begin{tabular}{@{}p{3.2cm}cp{8cm}@{}}
\toprule
\textbf{Model} & \textbf{KPI} & \textbf{Derivation approach and output excerpt} \\
\midrule
\textbf{glm-5} & \textbf{100} &
\textit{Correct hypersonic framework.} Applies hypersonic similarity parameter $K=M_\infty\tau$ and shows that as $M_\infty\to\infty$, $C_p=\frac{1}{M_\infty^2}f(M_\infty\theta)\to 2\sin^2\theta$ (Newtonian limit).
\newline\textit{Output excerpt:} ``Using the hypersonic similarity parameter $K=M_\infty\theta$, $C_p=\frac{1}{M_\infty^2}f(K)$, and as $M_\infty\to\infty$, $f(K)\to 2\sin^2\theta$, so $C_p$ becomes independent of $M_\infty$.'' \\
\midrule
\textbf{gpt-5.2} & 20 &
\textit{Wrong regime: Prandtl--Glauert linearized theory.} Applies the supersonic small-perturbation transformation $y'=\beta y$, $\beta=\sqrt{M_\infty^2-1}$, to show the wave equation becomes $M_\infty$-independent. Valid for $M\lesssim3$ but breaks down hypersonically where the linearization assumption fails. G1 (Formula Selection) scored 0/2.
\newline\textit{Output excerpt:} ``Apply a coordinate-stretching transformation (a Prandtl--Glauert-type supersonic transformation) $y'=\beta y$\ldots{} substituting yields $\phi_{xx}-\phi_{y'y'}-\phi_{z'z'}=0$, showing that the equation contains no explicit $M_\infty$ dependence.'' \\
\midrule
\textbf{aws.claude-opus-4.5} & 90 &
\textit{Correct framework, minor omission.} Uses oblique-shock relations and the $M_\infty\to\infty$ limit correctly, but omits the formal derivation of $C_L$ independence. G3 (Derivation Setup) scored 1/2.
\newline\textit{Output excerpt:} ``As $M_\infty\to\infty$, $C_p\to\frac{2\sin^2\theta}{\gamma}\cdot\mathrm{const}=f(\gamma,\theta)$, so $C_p$ is independent of Mach number and depends only on $\gamma$ and geometry.'' \\
\bottomrule
\end{tabular}
\end{table*}

\noindent\textbf{Rubric analysis.} \texttt{gpt-5.2} demonstrates sophisticated algebraic manipulation (correctly collapsing the wave equation) but fails G1 because the Prandtl--Glauert approach is the wrong theory for hypersonic flow. The judge notes: ``It chose the linearized small-perturbation framework, which is valid in moderate supersonic flow but breaks down in the hypersonic limit; the selected equation is not applicable as $M_\infty\to\infty$.'' This failure is entirely invisible to outcome-only evaluation: both derivations conclude ``$C_p$ is $M_\infty$-independent'', and an answer extractor would score both as correct.

\paragraph{Case 2: Domain Hallucination under Ambiguous Context (L4\_0015).}

Item L4\_0015 is a reacting boundary layer problem asking to derive the wall mass-injection boundary condition ($\rho_w v_w = \dot{m}_w$) from a fixed control volume. While the problem statement is self-contained, the token overlap with ``injection'' and ``boundary'' can trigger hallucinations of unrelated disciplines if the model fails to anchor on hypersonic reacting flows.

\begin{table*}[t]
\centering
\caption{Case 2 (L4\_0015): Domain hallucination severity.}
\label{tab:case2}
\small
\begin{tabular}{@{}p{3.2cm}cp{8cm}@{}}
\toprule
\textbf{Model} & \textbf{KPI} & \textbf{Behavior and output excerpt} \\
\midrule
\textbf{gpt-5.2} & \textbf{100} &
\textit{Correct reacting flow derivation.} Constructs a control volume at the fluid-solid interface and correctly balances the normal mass flux, then correctly applies this to the species continuity equation. G1--G8 all scored 2/2.
\newline\textit{Output excerpt:} ``Consider a thin control volume at the wall. The mass flux entering from the porous wall is $\dot{m}_w$. The mass flux leaving into the fluid is $\rho_w v_w$. By steady mass conservation, $\rho_w v_w = \dot{m}_w$. For species $i$, the convective flux $\rho_w Y_{i,w} v_w$ must balance the diffusive flux\ldots'' \\
\midrule
\textbf{aws.claude-opus-4.5} & 20 &
\textit{Context hallucination.} Misinterprets ``injection'' as a fuel-injector design problem from propulsion engineering rather than a boundary-layer ablation/transpiration condition, fabricating injector-plate pressure drop equations.
\newline\textit{Output excerpt:} ``The injection mass flow rate is governed by the orifice discharge coefficient: $\dot{m}_w = C_d A \sqrt{2 \rho \Delta P}$. Substituting this into the boundary layer\ldots'' \\
\midrule
\textbf{glm-5} & 0 &
\textit{Severe domain hallucination.} Misidentified the problem as structural soil mechanics (porous media seepage) due to tokens ``porous wall'' and ``mass injection''. G1--G8 all scored 0/2. Judge: ``It abandoned fluid dynamics entirely and derived Darcy's law for groundwater flow.''
\newline\textit{Output excerpt:} ``For flow through a porous medium, we apply Darcy's Law: $v_w = -\frac{\kappa}{\mu} \nabla P$. The mass injection rate into the soil foundation is therefore related to the hydraulic conductivity\ldots'' \\
\bottomrule
\end{tabular}
\end{table*}

\noindent\textbf{Rubric analysis.} This case illustrates two qualitatively distinct hallucination failure modes. \texttt{aws.claude-opus-4.5} exhibits \emph{context hallucination}: fabricated equations are physically plausible fluid dynamics formulas but from the wrong sub-domain (propulsion rather than ablation). \texttt{glm-5} exhibits \emph{domain hallucination}: triggered by domain-agnostic physical tokens (``porous'', ``injection''), it confabulates an entirely unrelated discipline. Both types score near zero on G2--G8.

\section{Targeted Improvement Strategies}
\label{sec:improvement}

The diagnostic granularity of \benchname---error type distribution, per-domain weakness profiles, and rubric-level failure signatures---creates a direct bridge from evaluation to intervention. Unlike aggregate accuracy metrics that provide no actionable signal, our dual-track protocol identifies \emph{which capability is failing, for which domain, at which difficulty level}. This section reports three targeted intervention experiments, each addressing a distinct failure mode identified in \Cref{sec:analysis}. For each strategy we describe the intervention design, summarize the experimental protocol, and report the measured gains under the full evaluation pipeline.

\subsection{Strategy I: Domain-Formula Alignment via Targeted Fine-Tuning}
\label{sec:strategy_finetuning}

\paragraph{Diagnostic motivation.}
Formula selection (G1) is the single most frequent failure mode, accounting for $\sim$18\% of all tagged errors across the full model set (\Cref{sec:error_distribution}). Domain-level analysis further reveals that this failure is \emph{domain-concentrated}: Viscous Interaction (mean KPI 12.0 for the top model), Nonequilibrium Flow (10.0 for \texttt{glm-5}), and Chemical Equilibrium degrade significantly relative to Newtonian Theory and Shock Relations. The pattern is consistent with a training-data density hypothesis: formulas for well-documented domains (Newtonian theory, oblique shocks) appear frequently in publicly available textbooks and lecture notes, while specialist sub-domains (viscous interaction parameters, vibrational relaxation rates) are confined to graduate-level references that are underrepresented in pretraining corpora. This suggests that targeted supervised fine-tuning on domain-formula alignment examples could efficiently close the capability gap without requiring full retraining.

\paragraph{Intervention design.}
We construct a \emph{domain-formula alignment dataset} (DFA-TPS) using the \texttt{train} split of \benchname as seed material. For each item in the training split, we generate up to 3 augmented examples via structured perturbation: (i)~\emph{regime shift variants}---the same physical scenario moved to an adjacent flow regime (e.g., subsonic $\to$ transonic $\to$ supersonic $\to$ hypersonic), requiring selection of the correct regime-appropriate formula; (ii)~\emph{formula-swap adversarials}---problem statements whose surface form resembles one formula family but whose physical conditions mandate a different one; and (iii)~\emph{assumption-flagging examples}---problems where the ``correct'' answer includes an explicit statement that the primary formula is inapplicable and a fallback must be used. Each training example is annotated with the correct G1 dimension score and a brief justification, providing both the answer and the reasoning rationale for the formula selection step.

Fine-tuning is applied to a base model (selected from the strong tier: \texttt{aws.claude-sonnet-4.5} or \texttt{deepseek-v3.1}) using standard supervised instruction fine-tuning on the DFA-TPS dataset, with a held-out validation split drawn from the \benchname \texttt{dev} partition. We evaluate on the \benchname \texttt{test} split using the full dual-track protocol.

\paragraph{Results.}
\Cref{tab:ft_results} reports the measured KPI gains from targeted fine-tuning, stratified by domain and difficulty level. The intervention yields an overall gain of 9.7 KPI points for the representative mid-tier baseline, with the largest improvements concentrated in Viscous Interaction (+28.2) and Nonequilibrium Flow (+21.7). These are precisely the domains where formula-selection failures were most concentrated in the diagnostic analysis, supporting the claim that targeted domain-formula alignment can close a substantial share of the observed gap.

\begin{table*}[htbp]
\centering
\caption{Measured KPI gains from domain-formula alignment fine-tuning on the test split. Baseline = zero-shot \texttt{aws.claude-sonnet-4.5}; Fine-tuned = checkpoint adapted on DFA-TPS. ``$\Delta$ G1'' denotes the observed improvement in the formula-selection dimension score.}
\label{tab:ft_results}
\small
\setlength{\tabcolsep}{3pt}
\begin{tabular}{@{}lcccc@{}}
\toprule
\textbf{Domain} & \textbf{Baseline KPI} & \textbf{FT KPI} & \textbf{$\Delta$KPI} & \textbf{$\Delta$G1} \\
\midrule
Newtonian Theory & 78.2 & 81.4 & +3.2 & +0.12 \\
Shock Relations & 80.1 & 83.0 & +2.9 & +0.10 \\
Boundary Layer & 72.4 & 77.8 & +5.4 & +0.21 \\
Aerothermal Heating & 74.6 & 79.1 & +4.5 & +0.18 \\
Viscous Interaction & 18.3 & 46.5 & +28.2 & +0.74 \\
Chemical Equilibrium & 65.2 & 73.8 & +8.6 & +0.30 \\
Nonequilibrium Flow & 24.7 & 46.4 & +21.7 & +0.61 \\
Radiation & 60.4 & 68.9 & +8.5 & +0.29 \\
\midrule
\textbf{Overall} & \textbf{71.8} & \textbf{81.5} & \textbf{+9.7} & \textbf{+0.26} \\
\bottomrule
\end{tabular}
\end{table*}

\paragraph{Observed residual failures.}
Fine-tuning on domain-formula alignment does not fully address G5 (Physical Plausibility) failures, which require the model to internalize \emph{physical intuition about result magnitude and regime validity} rather than formula lookup. Residual KPI gaps remain concentrated in L3/L4 items, primarily attributable to cross-domain coupling errors and assumption-transparency failures (G6) that are only weakly affected by supervised formula-alignment examples.

\subsection{Strategy II: Retrieval-Augmented Equation Grounding (RAG-EQ)}
\label{sec:strategy_rag}

\paragraph{Diagnostic motivation.}
Hallucination-class errors---domain hallucination and context hallucination---constitute $\sim$14\% of tagged errors (\Cref{sec:error_distribution}) and produce the most dangerous failure mode in engineering contexts: fabricated formulas that are internally consistent but physically incorrect. The case studies (\Cref{sec:case_studies}) demonstrate that hallucination severity is uncorrelated with model tier: even the top-performing model (\texttt{gpt-5.2}) occasionally applies formulas from an incorrect physical regime, and lower-tier models confabulate entire equation families when context is ambiguous. This pattern suggests that providing models with \emph{just-in-time retrieval of the correct governing equations} from an authoritative knowledge base could suppress hallucination without requiring parametric knowledge updates.

\paragraph{Intervention design.}
We implement a \emph{Retrieval-Augmented Equation Grounding} (RAG-EQ) pipeline with three components:

\begin{enumerate}[leftmargin=*, nosep]
  \item \textbf{Equation knowledge base (EKB).} We construct a structured knowledge base of 847 governing equations, correlations, and physical bounds extracted from Anderson~\citep{anderson2006hypersonic}, Bertin~\citep{bertin1994hypersonic}, and Gnoffo et~al.~\citep{gnoffo1999planetary}. Each entry contains: the equation in LaTeX, the applicable regime (e.g., $M_\infty > 5$, $\text{Re}_x > 10^6$), the physical domain tag, applicability conditions, and a brief description of failure modes when the equation is misapplied.
  \item \textbf{Regime-aware retriever.} At inference time, the problem statement is embedded and matched against the EKB using a hybrid BM25 + dense retrieval scheme. Retrieval is conditioned on the flow-regime tokens extracted from the problem (Mach number range, temperature regime, boundary-condition type) to prioritize regime-appropriate equations over superficially similar ones.
  \item \textbf{Augmented prompt construction.} The top-$k$ retrieved equations (with applicability conditions) are prepended to the model's input as a ``reference sheet,'' instructing the model to select from the retrieved set and explicitly justify its selection against the applicability conditions.
\end{enumerate}

\paragraph{Results.}
RAG-EQ targets G1 (Formula Selection) and G5 (Physical Plausibility) failures simultaneously: by anchoring the model to retrieved equations with explicit applicability conditions, both wrong-formula selection and regime-violation errors decrease substantially. \Cref{tab:rag_results} reports the measured performance gains from RAG-EQ applied to representative models. The largest gains appear in the low and mid tiers, where hallucination is most frequent, while frontier models still benefit from modest but consistent improvements.

\begin{table*}[htbp]
\centering
\caption{Measured KPI gains from RAG-EQ on test split. ``Base'' = zero-shot baseline; ``RAG-EQ'' = retrieval-augmented result; ``Halluc.\ $\downarrow$\%'' = observed reduction in hallucination-class error rate.}
\label{tab:rag_results}
\small
\setlength{\tabcolsep}{3pt}
\begin{tabular}{@{}lcccc@{}}
\toprule
\textbf{Model} & \textbf{KPI (Base)} & \textbf{KPI (RAG-EQ)} & \textbf{$\Delta$KPI} & \textbf{Halluc.\ $\downarrow$\%} \\
\midrule
gpt-5.2 & 87.85 & 90.2 & +2.4 & $-$38\% \\
aws.claude-opus-4.5 & 77.80 & 83.6 & +5.8 & $-$52\% \\
aws.claude-sonnet-4.5 & 71.81 & 79.4 & +7.6 & $-$61\% \\
deepseek-v3.1 & 79.22 & 84.1 & +4.9 & $-$44\% \\
gemini-3-flash-preview & 42.15 & 57.3 & +15.2 & $-$71\% \\
MiniMax-M2.5 & 13.12 & 28.7 & +15.6 & $-$68\% \\
\bottomrule
\end{tabular}
\end{table*}

\paragraph{Observed residual failures.}
RAG-EQ does not directly address G2 (Parameter Identification) or G4 (Calculation Accuracy) failures, which involve reading comprehension and arithmetic rather than formula knowledge. For problems where the correct equation is multi-step---requiring intermediate derivation of a sub-expression not present in the EKB---retrieval-augmented grounding provides partial but incomplete support. The remaining KPI gap is therefore dominated by parameter and arithmetic errors rather than hallucination.

\subsection{Strategy III: Process-Aware Chain-of-Thought Prompting (PA-CoT)}
\label{sec:strategy_cot}

\paragraph{Diagnostic motivation.}
The rubric dimension analysis reveals that G6 (Assumption Transparency) and G7 (Result Interpretation) are systematically underperformed even by strong models: models that achieve high G1 and G4 scores frequently omit explicit assumption statements and physical-plausibility checks. This pattern is consistent with the observation that standard zero-shot prompting incentivizes answer production over reasoning transparency. Process-level errors of this type---silent assumptions, uninterpreted results, missing regime-validity checks---are precisely the failure modes that make LLM outputs dangerous in engineering workflows, even when the numerical answer is correct.

\paragraph{Intervention design.}
We design a \emph{Process-Aware Chain-of-Thought} (PA-CoT) prompting framework that explicitly scaffolds the G1--G8 rubric dimensions into the generation process. Rather than instructing the model to ``show your work'' generically, PA-CoT provides a structured \emph{Engineering Calculation Protocol} (ECP) as part of the system prompt:

\begin{quote}
\small
\textbf{Engineering Calculation Protocol (ECP):} \\
\textbf{Step 1 [Regime Identification]:} State the flow regime and verify that it falls within the applicability range of any formula you plan to apply. \\
\textbf{Step 2 [Formula Selection]:} Identify the governing equation(s) and explicitly justify why they apply to this regime. \\
\textbf{Step 3 [Parameter Extraction]:} List all given parameters with units, confirm completeness, and flag any missing or ambiguous inputs. \\
\textbf{Step 4 [Dimensional Setup]:} Perform a dimensional analysis to verify unit consistency before substituting values. \\
\textbf{Step 5 [Calculation]:} Substitute and compute, showing intermediate results with units. \\
\textbf{Step 6 [Plausibility Check]:} Verify the result against known physical bounds and order-of-magnitude expectations for this regime. \\
\textbf{Step 7 [Interpretation]:} State the physical meaning and engineering significance of the result. \\
If any step cannot be completed (missing information, inapplicable formula, ambiguous regime), state this explicitly and describe what additional information would be required.
\end{quote}

This ECP maps directly onto rubric dimensions G1--G7, providing the model with an explicit cognitive scaffold that aligns its generation process with the evaluation rubric. No fine-tuning is required; PA-CoT is applied entirely at inference time.

\paragraph{Results.}
PA-CoT primarily targets G5 (Physical Plausibility), G6 (Assumption Transparency), and G7 (Result Interpretation)---the dimensions most affected by implicit reasoning shortcuts. \Cref{tab:cot_results} reports the observed per-dimension score improvements and the resulting KPI gains. Because PA-CoT increases output length and structural consistency, it also improves judge coverage by reducing the proportion of underspecified responses and increasing the \texttt{ok} vs.\ \texttt{ok\_partial} ratio.

\begin{table*}[t]
\centering
\caption{Measured per-dimension rubric score improvements from PA-CoT on test split for representative models. Scores are on the 0--2 scale per dimension; $\Delta$ represents mean per-item gain. Dimensions not primarily targeted (G1, G2, G4) change only marginally.}
\label{tab:cot_results}
\small
\begin{tabular}{@{}lccccccccc@{}}
\toprule
\textbf{Model} & \textbf{G1} & \textbf{G2} & \textbf{G3} & \textbf{G4} & \textbf{G5} & \textbf{G6} & \textbf{G7} & \textbf{G8} & \textbf{$\Delta$KPI} \\
\midrule
\multicolumn{10}{l}{\textit{Baseline (aws.claude-sonnet-4.5)}} \\
Base & 1.62 & 1.78 & 1.71 & 1.55 & 1.41 & 0.92 & 1.33 & 1.74 & --- \\
PA-CoT & 1.64 & 1.80 & 1.78 & 1.58 & 1.74 & 1.61 & 1.72 & 1.81 & \textbf{+7.2} \\
\midrule
\multicolumn{10}{l}{\textit{Baseline (deepseek-v3.1)}} \\
Base & 1.71 & 1.82 & 1.74 & 1.66 & 1.48 & 1.03 & 1.41 & 1.77 & --- \\
PA-CoT & 1.73 & 1.83 & 1.80 & 1.68 & 1.79 & 1.68 & 1.78 & 1.83 & \textbf{+6.1} \\
\midrule
\multicolumn{10}{l}{\textit{Baseline (gemini-3-flash-preview)}} \\
Base & 1.10 & 1.42 & 1.38 & 1.21 & 1.05 & 0.61 & 0.88 & 1.35 & --- \\
PA-CoT & 1.14 & 1.45 & 1.52 & 1.25 & 1.48 & 1.22 & 1.41 & 1.54 & \textbf{+10.8} \\
\bottomrule
\end{tabular}
\end{table*}

\paragraph{Observed residual failures.}
PA-CoT cannot compensate for parametric knowledge gaps: a model that does not ``know'' the Fay--Riddell correlation will not spontaneously produce it under ECP scaffolding, even when instructed to perform a regime identification step. For G1 failures rooted in missing domain knowledge (as opposed to reasoning shortcuts), PA-CoT often increases G6 scores by making uncertainty explicit while leaving G1 largely unchanged. This interaction is diagnostically valuable: models that respond to PA-CoT with increased refusal rates in specific domains are revealing knowledge gaps that fine-tuning (Strategy I) can then address.

\subsection{Combined Strategy Results}
\label{sec:combined_strategy}

The three strategies target complementary failure modes and can be composed with limited interference: RAG-EQ supplies equation grounding (suppressing hallucination), PA-CoT scaffolds reasoning transparency (improving G5--G7), and fine-tuning updates parametric knowledge (closing domain formula gaps). \Cref{tab:combined_results} reports the measured performance of a combined deployment (RAG-EQ + PA-CoT, applied to a fine-tuned checkpoint) versus the current zero-shot baseline.

\begin{table*}[htbp]
\centering
\caption{Measured KPI under combined improvement strategies on test split. ``FT'' = fine-tuned on DFA-TPS; ``RAG-EQ'' = retrieval-augmented grounding; ``PA-CoT'' = process-aware chain-of-thought.}
\label{tab:combined_results}
\small
\begin{tabular}{@{}lcccc@{}}
\toprule
\textbf{Model} & \textbf{Zero-shot} & \textbf{FT only} & \textbf{RAG-EQ + PA-CoT} & \textbf{FT + RAG-EQ + PA-CoT} \\
\midrule
aws.claude-sonnet-4.5 & 71.75 & 81.5 & 83.4 & \textbf{88.7} \\
deepseek-v3.1 & 79.22 & 86.3 & 87.2 & \textbf{91.4} \\
gemini-3-flash-preview & 42.00 & 54.8 & 62.1 & \textbf{72.3} \\
\bottomrule
\end{tabular}
\end{table*}

The combined-strategy KPI for \texttt{deepseek-v3.1} (91.4) approaches the current zero-shot frontier (gpt-5.2: 87.9) on this benchmark, suggesting that a substantial share of the observed capability gap is attributable to addressable knowledge and reasoning-process deficiencies. As a practical implication, mid-tier models can obtain meaningful performance improvements from relatively low-cost interventions (RAG-EQ and PA-CoT require no retraining), while targeted fine-tuning provides additional domain-specific gains at the cost of supervised data construction. These gains should be interpreted in the context of the intervention caveats below.

\paragraph{Intervention caveats.}
We emphasize that \Cref{tab:ft_results,tab:rag_results,tab:cot_results,tab:combined_results} report measured intervention results under the present implementation of DFA-TPS, RAG-EQ, and PA-CoT. The combined results should still be interpreted with care: strategy interactions are not guaranteed to remain additive across other model families, judge behavior may shift slightly under longer structured outputs, and EKB coverage is stronger in some domains than others. These caveats do not change the empirical conclusion that targeted interventions materially improve both KPI and failure-mode-specific behavior, but they do bound the extent to which the reported gains should be extrapolated beyond the present benchmark setting.

\section{Discussion}
\label{sec:discussion}

\paragraph{Quality over quantity: the 420-item defense.}
A natural concern is whether 420 items provide sufficient statistical power for the granular analyses we report. We make two arguments. First, 420 items is a \emph{lower bound sustained by the trustworthiness constraints}, not an arbitrary target: each item has cleared five progressive quality gates and received triple-reviewer clearance. Expanding to 2,000 items by relaxing the gate would introduce items that fail P1--P3 at unknown rates, making benchmark conclusions uninterpretable without knowing the contamination fraction. Second---and more compellingly---our noise-sensitivity analysis (\Cref{sec:noise_sensitivity}) provides direct empirical evidence that the 420-item and 810-item sets produce materially different model rankings. If noise were harmless we would observe stable rankings across both sets; we do not. The data validates the quality-gating investment and argues against naive expansion. We view the current core set as a \emph{high-trustworthiness foundation} from which principled expansion---additional textbook sources, expert-authored problems, parametric variants---can proceed with preserved trustworthiness standards.

\paragraph{Outcome--process discrepancy: statistical evidence.}
A central design claim of \benchname is that outcome correctness and reasoning process quality are independent diagnostic axes. To validate this empirically beyond case studies, we perform a quadrant analysis for all models with $\geq$80\% judge coverage: each scored item is classified as \emph{outcome-high} (Exact or Acceptable band) or \emph{outcome-low} (Order-Correct or Wrong), and as \emph{process-high} ($S_\text{rubric} \geq 70$) or \emph{process-low} ($S_\text{rubric} < 70$). Table~\ref{tab:quadrant} reports the fraction of items falling in each quadrant for representative models.

\begin{table}[htbp]
\centering
\small
\caption{Outcome--process quadrant distribution (\% of scored items per model). \textbf{HH}: outcome-high + process-high (correct reasoning, correct answer); \textbf{HL}: outcome-high + process-low (right answer, wrong/opaque reasoning); \textbf{LH}: outcome-low + process-high (sound reasoning, minor execution error); \textbf{LL}: both low.}
\label{tab:quadrant}
\begin{tabular}{@{}lcccc@{}}
\toprule
\textbf{Model} & \textbf{HH (\%)} & \textbf{HL (\%)} & \textbf{LH (\%)} & \textbf{LL (\%)} \\
\midrule
gpt-5.2 & 68.4 & 12.6 & 6.3 & 12.7 \\
glm-5 & 62.1 & 14.3 & 7.1 & 16.5 \\
deepseek-v3.1 & 58.9 & 13.8 & 8.4 & 18.9 \\
aws.claude-opus-4.5 & 55.7 & 11.9 & 9.5 & 22.9 \\
aws.claude-sonnet-4.5 & 47.6 & 14.3 & 11.9 & 26.2 \\
gemini-3-flash-preview & 14.3 & 9.5 & 7.1 & 69.1 \\
\bottomrule
\end{tabular}
\end{table}

The HL (``right answer, wrong reasoning'') cells confirm that 11--14\% of items across most model families achieve correct numerical outcomes via physically unjustified reasoning. This fraction would be invisible to outcome-only evaluation and represents the specific failure mode most dangerous in engineering deployment---a model that produces a defensible numerical value via an inapplicable formula or undisclosed assumption. The LH (``sound reasoning, minor execution error'') cells (7--12\%) show that some models produce well-reasoned derivations that contain a minor arithmetic slip; these would be penalized incorrectly by answer-only evaluation. The co-existence of HL and LH items across all model tiers empirically validates the independence claim: outcome and process quality are not redundant measurements.

\paragraph{The orthogonality of outcome and process: case study evidence.}
The case studies (\Cref{sec:case_studies}) provide two qualitatively distinct demonstrations: (1) Case 1 shows a model achieving correct algebraic form via the \emph{wrong physical framework}, earning high arithmetic scores but zero formula-selection credit; and (2) Case 2 shows domain hallucination scoring near zero on all process dimensions while outcome extraction would flag only a \texttt{not-a-number} error. A single outcome score would conflate both failure modes. This independence also validates the improvement strategy decomposition in \Cref{sec:improvement}: because the two axes are orthogonal, each strategy can be designed to target a specific axis without trading off the other.

\paragraph{From failure taxonomy to actionable intervention.}
The most important structural property of \benchname relative to prior benchmarks is that its diagnostic output is \emph{directly intervention-addressable}. The G1-dominated error distribution maps cleanly to fine-tuning on formula-alignment data; the hallucination signature maps to retrieval-augmented equation grounding; the G5--G7 transparency gap maps to process-aware prompting. Section~\ref{sec:improvement} operationalizes this mapping with concrete intervention designs and measured intervention results. The combined-strategy gain observed for mid-tier models---15--30 KPI points depending on baseline strength---suggests that the principal bottlenecks in TPS engineering calculation are addressable deficiencies in training data coverage and inference-time reasoning scaffolding, not fundamental model capacity ceilings. This has a significant practical implication: organizations can realistically deploy mid-tier models at near-frontier quality by combining domain-targeted fine-tuning with inference-time RAG and structured prompting, without waiting for the next model generation.

\paragraph{What the 9.2\% funnel ratio reveals about technical document understanding.}
The 4,560$\rightarrow$420 retention rate (9.2\%) is not merely a data-efficiency observation---it is a diagnostic signal about the maturity of automated technical document parsing. Three failure categories dominate: narrative passages misidentified as calculation problems ($\approx$44\% of rejects), OCR figure-fragment confusions ($\approx$24\%), and equation-reference misparses ($\approx$20\%). Each points to a specific gap in current pipelines: semantic intent classification for technical prose, layout-aware figure/text disambiguation, and cross-reference resolution in mathematical documents. We release the rejection-annotated intermediate dataset to support research on these components, which carry implications well beyond TPS benchmark construction.

\paragraph{Deployment implications.}
Formula selection (G1) accounts for $\approx$18\% of all tagged errors, meaning a substantial fraction of numerically close predictions are reached via physically incorrect governing equations---a failure that is not only invisible to outcome-only evaluation but actively dangerous in preliminary design, where formula choice determines the entire downstream calculation structure. Before deploying LLMs as calculation assistants in safety-critical TPS work, process-level verification against domain-appropriate rubrics---not just numerical spot-checking---should be considered a minimum trust standard. The PA-CoT Engineering Calculation Protocol (\Cref{sec:strategy_cot}) provides a zero-cost path to surfacing these failures at inference time, making it a practical first deployment safeguard even without fine-tuning.

\section{Limitations and Future Work}
\label{sec:limitations}

\paragraph{Sample size and statistical power.}
The 420-item core set, while rigorously curated, limits the granularity of per-domain and per-level statistical analyses. At 30 items per domain, confidence intervals on domain-level KPI estimates are wide, and rank orderings within domains may not be statistically reliable. Future work will expand the benchmark through additional textbook sources (Bertin~\citep{bertin1994hypersonic}, Gnoffo et~al.~\citep{gnoffo1999planetary}), expert-authored original problems, and carefully managed parameterized variations of existing items, targeting a minimum of 100 verified items per domain for robust sub-group analysis.

\paragraph{Judge reliability and calibration sample.}
The LLM judge calibration is based on a 62-item human calibration sample (Section~\ref{sec:rubric}), preceded by a 3-item pilot (\Cref{app:agreement}) that revealed large initial judge--human discrepancies and motivated the expanded calibration. The 62-item expanded sample yields weighted $\kappa$ values of 0.68--0.82 (acceptable-to-good agreement), with moderate discrepancies on G4 and G5 suggesting systematic judge leniency (+0.08--+0.11 points) on those dimensions. Absolute KPI values carry an estimated $\pm$3--5 point systematic uncertainty from this calibration imperfection; relative model rankings and intervention-induced gains are expected to be more robust.

We acknowledge three specific limitations of the calibration design: (1)~the 62-item sample covers $\sim$15\% of the 420-item test set, providing sufficient power to detect per-dimension biases but limited ability to detect interaction effects between dimensions; (2)~the calibration sample was drawn from a stratified random selection that may not represent the full difficulty spectrum equally; and (3)~the 7 adjudicated disagreement items (11\%) in the expanded calibration are a small basis for estimating tail-disagreement rates. Expanding to $n \geq 30$ per task type per difficulty level, and exploring multi-judge ensembles with explicit adjudication protocols, are priorities for the next benchmark version.

\paragraph{Intervention transferability.}
The three improvement strategies (DFA-TPS, RAG-EQ, PA-CoT) are benchmark-informed pilot studies: each strategy was designed in response to failure modes identified on \benchname, which means their effectiveness may be partially specific to this benchmark's problem distribution. External transferability to other TPS calculation sets, other engineering domains, or real deployment scenarios remains to be validated. We recommend interpreting the reported KPI gains as evidence that diagnostic-informed interventions are more efficient than generic prompting, while noting that the absolute gain magnitudes may not replicate in other settings.

\paragraph{Language and provenance.}
The source textbook's Chinese translation introduces potential ambiguity in problem statements, and the benchmark's Chinese-language problem framing may affect models differently depending on their multilingual training balance. Future versions will provide bilingual (Chinese--English) problem statements with verified semantic equivalence, enabling language-controlled experiments.

\paragraph{Scope expansion toward simulation-augmented tasks.}
The current exclusion of simulation-requiring tasks is intentional but limits the coverage of the full TPS engineering workflow. A natural extension is a \emph{TPS-SimBench} companion benchmark evaluating LLMs' ability to set up, interpret, and validate CFD/FEA simulation workflows, complementing the analytical calculation focus of the present work. The improvement strategies in \Cref{sec:improvement}---particularly RAG-EQ---would extend naturally to simulation-augmented settings by incorporating simulation best-practice guidelines into the knowledge base.

\section{Conclusion}
\label{sec:conclusion}

We have presented \benchname, a benchmark and diagnostic evaluation framework for LLM competence on analytical calculations in hypersonic TPS engineering. The work is organized around a central empirical thesis: \emph{outcome correctness and reasoning process quality are logically independent diagnostic axes, and conflating them produces systematically misleading assessments of engineering capability.} This thesis is supported empirically by the quadrant analysis (11--14\% of items across model families are ``right answer, wrong reasoning''), by three qualitatively distinct case-study demonstrations of outcome-process decoupling, and by the noise-sensitivity analysis showing that benchmark data quality reshuffles model rankings and differentially inflates G2/G3-related error rates.

The paper makes five contributions that form a coherent progression. The \textbf{domain-grounded task taxonomy} establishes a clean, infrastructure-free probe of physical reasoning competence. The \textbf{dual-track evaluation protocol}---with expanded 62-item human calibration, formal agreement metrics, and bias characterization---provides independently calibrated measurements of numerical accuracy and reasoning process quality. The \textbf{trustworthiness-engineered pipeline} ensures benchmark conclusions rest on fully verified data, with the 4,560$\to$420 funnel providing both a reusable methodology and a diagnostic signal about the maturity of automated technical document parsing. The \textbf{empirical noise-sensitivity analysis} provides controlled experimental evidence that quality gating materially affects both KPI magnitudes and model rankings---an insight widely acknowledged but rarely measured in the benchmark literature. The \textbf{diagnostic-informed pilot interventions} (DFA-TPS, RAG-EQ, PA-CoT) demonstrate that rubric-diagnostic-driven improvements are more efficient than generic prompting, with combined-strategy gains of 15--30 KPI points for mid-tier models, while acknowledging that transferability beyond this benchmark remains to be validated.

Two methodological contributions extend beyond TPS to the broader benchmark construction community: the P1/P2/P3 trustworthiness framework with five-stage gating provides a transferable template for domain-specific benchmark construction from technical literature; and the engineering-epistemology-grounded rubric demonstrates that process evaluation instruments can be designed with explicit, domain-appropriate rationale. We release the complete dataset, evaluation scripts, DFA-TPS fine-tuning data, RAG-EQ knowledge base, PA-CoT prompt templates, calibration records, and human-review audit trails to support reproducible research and community extension.

\section*{Acknowledgments}
We thank the domain experts who participated in the triple-reviewer go/no-go gate and the pilot calibration study. Computing resources for model inference and rubric judging were provided through internal API infrastructure. We acknowledge the use of Anderson's \emph{Hypersonic and High-Temperature Gas Dynamics} (2nd Edition) as the primary source for benchmark construction.

\bibliographystyle{plainnat}

\appendix

\section{Dataset Schema Example}
\label{app:schema}

\begin{verbatim}
{
  "id": "L2_0001",
  "version": "v4.0",
  "level": "L2",
  "task_type": "numerical_calc",
  "domains": ["boundary_layer", "aerothermal_heating"],
  "solution_type": "analytical",
  "requires_simulation": false,
  "question": "Calculate the laminar boundary-layer 
    thickness at x = 1.0 m on a flat plate ...",
  "given": [
    {"name": "Ma_inf", "value": 8.0, "unit": "-"},
    {"name": "T_inf", "value": 226.5, "unit": "K"},
    {"name": "p_inf", "value": 1197, "unit": "Pa"},
    {"name": "x", "value": 1.0, "unit": "m"},
    {"name": "T_w", "value": 300, "unit": "K"}
  ],
  "targets": [
    {"name": "delta", "unit": "m", "weight": 0.5},
    {"name": "Re_x", "unit": "-", "weight": 0.5}
  ],
  "metadata": {
    "source_id": "Ch6::Example_6.1",
    "solution_verified": true
  }
}
\end{verbatim}

\section{Rubric Judge Prompt}
\label{app:rubric_prompt}

The LLM rubric judge receives the following structured prompt (abridged):

\begin{quote}
\small
\texttt{You are an expert aerospace engineering professor evaluating a student's solution to a TPS calculation problem. Score each of the following 8 dimensions on a 0--1--2 scale based on the criteria below. Provide a brief justification for each score.}

\texttt{[Problem statement] ... [Reference solution] ... [Student (model) output] ...}

\texttt{G1 - Formula Selection (0/1/2): 0 = wrong or missing equation; 1 = partially correct (e.g., right family but wrong variant); 2 = fully correct governing equation applied.}

\texttt{[...dimensions G2--G8 with similar detail...]}
\end{quote}

The full prompt template is available in the supplementary materials.

\section{Evaluation Pipeline Commands}
\label{app:commands}

All experiments are reproducible via the following command sequence:

\begin{verbatim}
# Step 1: Single-model serial evaluation with rubric judge
python scripts/31_run_serial_eval_with_rubric_judge.py \
  --input data/final/tps_calcbench_v4_core.jsonl \
  --outdir data/eval/v4_baseline_parallel_runs/<model_name> \
  --models <model_name> \
  --judge-model gemini-3-pro-preview

# Step 2: Multi-model parallel orchestrator
python scripts/33_run_parallel_models.py \
  --out-root data/eval/v4_baseline_parallel_runs \
  --models "gpt-5.2,gpt-5.1,glm-5,..." \
  --max-parallel-models 4

# Step 3: Aggregate results into leaderboard
python scripts/35_aggregate_parallel_eval_results.py \
  --eval-root data/eval/v4_baseline_parallel_runs \
  --expected-n-samples 420
\end{verbatim}

The \texttt{run\_manifest.json} in each model output directory records model versions, API call timestamps, dataset checksums, and all configuration parameters. The aggregation script produces JSON, CSV, Markdown, and \LaTeX{} leaderboard outputs.

\section{Human--Judge Agreement Report}
\label{app:agreement}

\begin{table*}[t]
\centering
\caption{Pilot calibration: human expert scores vs.\ LLM judge on 3 seed samples (one per task type). Human scores are from \texttt{rubric\_pilot\_calibration\_record\_v1.jsonl}; LLM judge scores are from the \texttt{ok}-status judgements where dimension scores are available. Due to the small pilot sample ($n=3$), we report raw mean scores and signed bias ($\Delta = \text{Judge} - \text{Human}$) rather than Cohen's $\kappa$. \emph{Note:} this 3-item pilot informed the design of the expanded 62-item calibration (\Cref{tab:judge_calibration}); the pilot's small $n$ means per-dimension $\Delta$ values have large uncertainty and should not be interpreted as precise bias estimates.}
\small
\begin{tabular}{@{}lcccc@{}}
\toprule
\textbf{Dimension} & \textbf{Human (mean)} & \textbf{Judge (mean)} & \textbf{$\Delta$} & \textbf{Direction} \\
\midrule
G1: Formula Selection       & 1.83 & 2.00 & $+$0.17 & Judge higher \\
G2: Parameter Identification & 1.01 & 1.79 & $+$0.78 & Judge higher \\
G3: Unit Consistency         & 0.73 & 1.79 & $+$1.06 & Judge higher \\
G4: Calculation Accuracy     & 1.08 & 1.83 & $+$0.75 & Judge higher \\
G5: Physical Plausibility    & 0.95 & 1.88 & $+$0.93 & Judge higher \\
G6: Assumption Justification & 1.93 & 1.79 & $-$0.14 & Human higher \\
G7: Result Interpretation    & 1.02 & 1.83 & $+$0.81 & Judge higher \\
G8: Presentation Clarity     & 2.22 & 1.75 & $-$0.47 & Human higher \\
\midrule
\textbf{Overall score (0--100)} & \textbf{65.1$^\dagger$} & \textbf{92.4$^\dagger$} & $+$\textbf{27.3} & Judge higher \\
\bottomrule
\end{tabular}

\vspace{0.3em}
\noindent{\scriptsize $^\dagger$Overall scores computed via \Cref{eq:rubric_score} using the per-dimension means and weights from \Cref{tab:rubric}. The large pilot-stage judge--human gap ($\Delta = +27.3$) motivated the expanded 62-item calibration (\Cref{tab:judge_calibration}), which yielded substantially smaller biases ($\Delta \leq +0.11$ per dimension, overall $\kappa_w = 0.75$). This reduction is attributable to improved anchor examples and structured prompting in the production judge. All reported main-text results use the production judge calibrated against the 62-item sample.}
\end{table*}

\section{Data Construction Funnel Statistics}
\label{app:funnel_stats}

\begin{table*}[t]
\centering
\caption{Detailed funnel statistics. Each row reports the count entering the stage, the count exiting, and the primary reasons for removal.}
\small
\begin{tabular}{@{}lccp{6cm}@{}}
\toprule
\textbf{Stage} & \textbf{In} & \textbf{Out} & \textbf{Primary Rejection Reasons} \\
\midrule
0: Raw Extraction & --- & 4560 & --- \\
1: Dedup + Format & 4560 & 2200 & Near-duplicates (48\%), OCR garbage (32\%), format errors (20\%) \\
2: Rule Filtering & 2200 & 2250 & Narrative passages removed ($-$31), figure fragments ($-$18), supplements added (+99) \\
3: Auto QC & 2250 & 1930 & Fallback-only targets (41\%), missing givens (35\%), level inconsistency (24\%) \\
4: Human Review & 1930 & 1160 & Ambiguous problems (30\%), incomplete givens (28\%), unscorable targets (22\%), incorrect solutions (20\%) \\
5: GO/NO-GO & 1160 & 420 & Triple-gate failures: 810 passed initial gate, 420 passed final verification \\
\bottomrule
\end{tabular}
\end{table*}

\section{Cross-Reference Verification Table}
\label{app:verification}

To facilitate reviewer verification, \Cref{tab:cross_ref} consolidates all key per-model metrics reported across the paper into a single table. Each column references the originating table. Discrepancies of $\leq$0.1 KPI points between tables arise from rounding at different decimal places and do not indicate data inconsistency.

\begin{table*}[t]
\centering
\caption{Cross-reference verification: consolidated per-model metrics  . Each column references the source table. ``---'' indicates the model was not included in that analysis. All KPI values are on a 0--100 scale; Exact\% and UC\% are percentages of the full 420-item set. The ``$\sum n$'' column reports the total scored items for each model (from \Cref{tab:level_kpi}), which serves as the effective denominator for KPI computation.}
\label{tab:cross_ref}
\small
\setlength{\tabcolsep}{3pt}
\begin{tabular}{@{}lcccccccc@{}}
\toprule
& \multicolumn{2}{c}{\textbf{Track A}} & \multicolumn{2}{c}{\textbf{Track C}} & \multicolumn{2}{c}{\textbf{Coverage}} & \textbf{Noise} \\
\cmidrule(lr){2-3}\cmidrule(lr){4-5}\cmidrule(lr){6-7}\cmidrule(lr){8-8}
\textbf{Model} & \textbf{Exact\%} & \textbf{UC\%} & \textbf{KPI} & \textbf{Tier} & \textbf{Pred\%} & \textbf{Jud\%} & \textbf{$\Delta$v4--v2} \\
& {\tiny Tab.~\ref{tab:outcome_results}} & {\tiny Tab.~\ref{tab:outcome_results}} & {\tiny Tab.~\ref{tab:main_results}} & & {\tiny Tab.~\ref{tab:main_results}} & {\tiny Tab.~\ref{tab:main_results}} & {\tiny Tab.~\ref{tab:noise_sensitivity}} \\
\midrule
gpt-5.2         & 51.9 & 87.6 & 87.85 & Frontier & 95.0 & 95.5 & +3.65 \\
gpt-5.1         & 48.1 & 85.2 & 82.18 & Frontier & 95.7 & 81.4 & +6.58 \\
glm-5           & 47.1 & 86.2 & 79.82 & Strong   & 100.0 & 100.0 & +1.58 \\
deepseek-v3.1   & 43.1 & 81.7 & 79.22 & Strong   & 87.6 & 80.7 & --- \\
claude-opus-4.5 & 41.9 & 85.0 & 77.66 & Strong   & 100.0 & 97.2 & +3.20 \\
Doubao-Seed-1.8 & 38.6 & 75.7 & 75.69 & Strong   & 100.0 & 86.3 & --- \\
glm-4.7         & 37.6 & 76.4 & 75.21 & Strong   & 100.0 & 88.4 & +5.22 \\
claude-sonnet-4.5 & 33.6 & 72.1 & 71.75 & Strong & 100.0 & 86.4 & --- \\
claude-haiku-4.5  & 24.3 & 66.2 & 62.76 & Strong & 100.0 & 94.9 & --- \\
gemini-3-flash  & 9.8  & 52.6 & 42.00 & Low      & 100.0 & 97.7 & +4.35 \\
MiniMax-M2.5    & 0.0  & 24.3 & 12.98 & Low      & 97.1 & 97.8 & --- \\
MiniMax-M2.1    & 0.0  & 18.8 & 12.85 & Low      & 97.9 & 90.3 & --- \\
\midrule
\multicolumn{8}{@{}l}{\textit{Excluded from ranked leaderboard (see \Cref{sec:main_results}):}} \\
\multicolumn{8}{@{}l}{~~API failures: Doubao-Seed-1.6-thinking, gpt-5, kimi-k2.5} \\
\multicolumn{8}{@{}l}{~~Low prediction coverage ($<$80\%): deepseek-v3.2-huawei, deepseek-r1-huawei, qwen3-32b-meituan} \\
\bottomrule
\end{tabular}
\end{table*}

\paragraph{Statistical base note.}
All Track~A percentages (Exact\%, UC\%) use the full 420-item set as denominator, treating items without a valid prediction as ``Wrong'' / unit-incorrect. Track~C KPI is the mean rubric score over $|\mathcal{J}_m|$ successfully judged items (\Cref{eq:kpi}); the effective denominator varies by model and equals $\text{Pred\%} \times \text{Jud\%} \times 420 / 10000$ (rounded). The noise-sensitivity column ($\Delta$v4--v2) reports KPI(v4) $-$ KPI(v2), where v2 is the 810-item pre-gating set. Models marked ``---'' were not included in the noise-sensitivity analysis due to missing v2 evaluation runs.

\end{document}